\documentclass[runningheads]{llncs}
\usepackage{graphicx}
\usepackage{tikz}
\usepackage{comment}
\usepackage{amsmath,amssymb} % define this before the line numbering.
\usepackage{color}
\usepackage{ulem}
\usepackage{xspace}
\usepackage{bm}
\usepackage{booktabs}
\usepackage{multirow}
\usepackage{enumitem}
\usepackage[multiple]{footmisc}

\definecolor{citec}{RGB}{21, 101, 192}
\definecolor{refc}{RGB}{220, 40, 40}
\usepackage[pagebackref=true,breaklinks=true,letterpaper=true,colorlinks,bookmarks=false,citecolor=citec,linkcolor=refc]{hyperref}

\let\llncssubparagraph\subparagraph
\let\subparagraph\paragraph
\usepackage[compact]{titlesec}
\let\subparagraph\llncssubparagraph

\usepackage[accsupp]{axessibility}  % Improves PDF readability for those with disabilities.

\usepackage[width=122mm,left=12mm,paperwidth=146mm,height=193mm,top=12mm,paperheight=217mm]{geometry}

\makeatletter
\newcommand{\printfnsymbol}[1]{%
        \textsuperscript{\@fnsymbol{#1}}%
}
\makeatother

\makeatletter
\DeclareRobustCommand\onedot{\futurelet\@let@token\@onedot}
\def\@onedot{\ifx\@let@token.\else.\null\fi\xspace}

\def\eg{\emph{e.g}\onedot} 
\def\ie{\emph{i.e}\onedot} 
 
\def\etc{\emph{etc}\onedot} 
 
\def\etal{\emph{et al}\onedot}
\makeatother

\begin{document}
% \renewcommand\thelinenumber{\color[rgb]{0.2,0.5,0.8}\normalfont\sffamily\scriptsize\arabic{linenumber}\color[rgb]{0,0,0}}
% \renewcommand\makeLineNumber {\hss\thelinenumber\ \hspace{6mm} \rlap{\hskip\textwidth\ \hspace{6.5mm}\thelinenumber}}
% \linenumbers
\pagestyle{headings}
\mainmatter
\def\ECCVSubNumber{****}  % Insert your submission number here

\title{NeuMesh: Learning Disentangled Neural Mesh-based Implicit Field \\ for Geometry and Texture Editing
}

% CAMERA READY SUBMISSION
\titlerunning{NeuMesh}
% If the paper title is too long for the running head, you can set
% an abbreviated paper title here

\author{
Bangbang Yang\inst{1}\thanks{Bangbang Yang and Chong Bao contributed equally to this work.} \and
Chong Bao\inst{1}\printfnsymbol{1} \and
Junyi Zeng\inst{1} \and
Hujun Bao\inst{1} \and
Yinda Zhang\inst{2}\printfnsymbol{2} \and
Zhaopeng Cui\inst{1}\thanks{Corresponding authors.} \and
Guofeng Zhang\inst{1}\printfnsymbol{2}
}

\authorrunning{Bangbang Yang and Chong Bao et al.}

\institute{
State Key Lab of CAD\&CG, Zhejiang University \and
Google
}

%******************
\maketitle

\begin{abstract}
Very recently neural implicit rendering techniques have been rapidly evolved and shown great advantages in novel view synthesis and 3D scene reconstruction.
However, existing neural rendering methods for editing purposes offer limited functionality, \eg, rigid transformation, or not applicable for fine-grained editing for general objects from daily lives.
In this paper, we present a novel mesh-based representation by encoding the neural implicit field with disentangled geometry and texture codes on mesh vertices, which facilitates a set of editing functionalities, including mesh-guided geometry editing, designated texture editing with texture swapping, filling and painting operations. To this end, we develop several techniques including learnable sign indicators to magnify spatial distinguishability of mesh-based representation, distillation and fine-tuning mechanism to make a steady convergence, and the spatial-aware optimization strategy to realize precise texture editing.
Extensive experiments and editing examples on both real and synthetic data demonstrate the superiority of our method on representation quality and editing ability. Code is available on the project webpage: \url{https://zju3dv.github.io/neumesh/}.


\end{abstract}

\section{Introduction}

Neural implicit field has achieved great success in 3D reconstruction and free-viewpoint rendering, and becomes a promising solution to take the place of traditional 3D shape and texture representation, \eg,  point cloud or textured mesh, due to its phenomenal rendering quality.
However, for 3D modeling and CG creation, artists still prefer to use mesh-based workflow across daily works.
For instance, in modern 3D CG software (\eg, Blender, Maya and 3ds Max), polygon mesh-based representations can be precisely controlled and edited, \ie, texturing with UV-map and changing shapes by altering vertices and faces, with all the previewed modification accurately reflected in the final rendering product.
Despite great progress made to improve the flexibility of the neural implicit field, including handling dynamic scenes~\cite{park2021nerfies,park2021hypernerf}, becoming scene agnostic~\cite{wang2021ibrnet,schwarz2020graf}, fast rendering~\cite{yu2021plenoxels,muller2022instant}, and scalability improvement~\cite{tancik2022block,xiangli2021citynerf}, the support of neural implicit field towards editing is still limited, \eg, on a very specific semantic category~\cite{deng2021deformed,liu2021editing,kania2021conerf} or purely rigid transformation~\cite{zhang2021editable,yang2021learning,guo2020object}. 
One plausible reason is that particular network encoding structures (\eg, coordinate-based MLP, voxels or scattered point cloud) are not compatible with fine-grained scene editing such as non-rigid geometry deforming and texture editing for a local region of interest, and thus cannot satisfy the broad demands of artistic creation.

\begin{figure*}[t!]
    \centering
    \includegraphics[width=1.0\linewidth, trim={0 0 0 0}, clip]{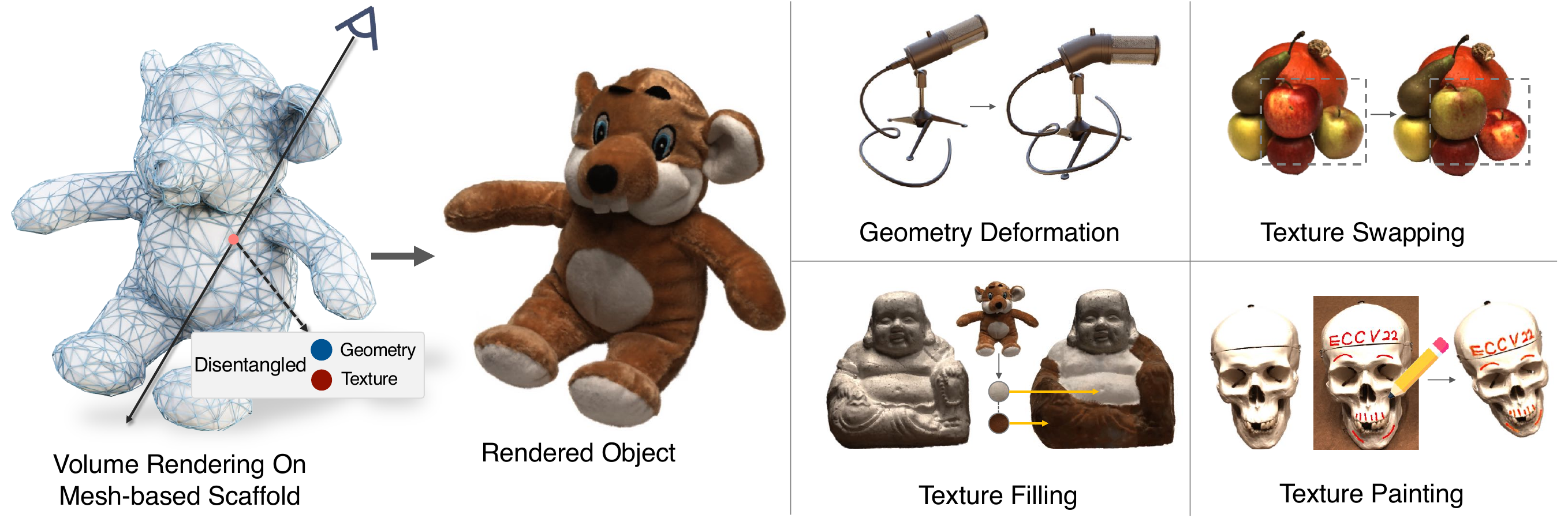}
    \caption{
    \textbf{NeuMesh.} We present a novel representation for volumetric neural rendering, which encodes the neural implicit field with disentangled geometry and texture features on a mesh scaffold. 
    With the locally separated latent codes, our representation enables a series of editing functionalities, including mesh-guided geometry deformation, designatable texture swapping, filling and painting.
    }
    \label{fig:teaser}
\end{figure*}

In this paper, we propose a novel neural implicit representation, NeuMesh, to facilitate editing in both 3D modeling and texturing. 
Our representation bares the following properties to seamlessly integrate with existing common workflow for 3D editing: 
\textbf{1)} The neural representation encodes scene with a series of vertex-bounded codes on a mesh scaffold and MLP-based decoders, instead of a pure MLP, point clouds or voxels, 
and can be deformed together with the mesh.
During the volume rendering, the implicit field is decoded via interpolation of these codes.
By doing so, any modification to the mesh geometry or local codes would precisely reflect the rendering output. 
\textbf{2)} The geometry and appearance representations are disentangled, \ie, encoded in two separate latent codes, such that texture can be transferred across geometry by replacing the appearance code from one another.
As shown in Fig.~\ref{fig:teaser}, our representation supports non-rigid object deformation with a handful approach (\eg, deforming with a mesh proxy), and provides various fashions of texture editing, including texture swapping of irregular mesh segments, texture filling at a specific area with pattern from a pre-captured object, and a user-friendly texture painting which reflects the philosophy of ‘what you get is what you see’.

However, learning and deploying such representation for rendering and editing is non-trivial.
\textbf{1)} Unlike voxel-based representation \cite{nsvf}, na\"ive trilinear code interpolation
is not sufficient to measure spatial variation since we dedicate to encoding the implicit field with a set of `single layer' codes on 
mesh vertices, and the inner/outer queries along the direction perpendicular to the surface lacks spatial distinguishability (\ie, failing to determine positive or negative direction when crossing through a mesh face).
A possible workaround is to complement the network input with signed distance to the mesh surface~\cite{neural_actor}, which, however, is not always available, especially on non-watertight/ill-defined geometries.
To tackle this challenge, we propose to maintain a set of learnable sign indicators for mesh vertices.
Then, for each query point along the ray, we compute a signed distance from nearby vertices by weighting the projected distances on the indicators.
In this way, our representation is completely agnostic to arbitrary mesh typologies (\eg, non-watertight or non-manifold meshes).
During the training process, these sign indicators are continuously adjusted to best fit the optimization objective.
\textbf{2)} Although such vertex-bounded and geometry-texture disentangled representation merits good flexibility on editing purpose, it does not preserve spatial continuity as MLP-based methods~\cite{nerf,neus,volsdf} and thus easily suffers from unstable training.
To mitigate this problem, we employ a distillation and fine-tuning training scheme, which leverages a pre-trained implicit field to guide the optimization of our representation.
In this way, we transfer a baked MLP-based implicit model into NeuMesh, the first neural rendering model that naturally inherits editable capability from the flexible mesh-based workflow.
\textbf{3)} To fulfill the demand for flexible and user-friendly texture editing operations (\eg, propagating 2D image painting to the 3D field), a na\"ive approach is fine-tuning with a single image.
However, this might let the network overfit to a specific view and the rendered images from other views degrade (\eg, introducing noticeable artifacts as shown in Fig.~\ref{fig:ablation_vis} (b)).
In order to solve this challenge, we propose a spatial-aware optimization strategy that is naturally derived from our representation, in which we select the affected texture codes with several probing rays from painted pixels to the mesh surface, and only fine-tuning these codes during the optimization.
Therefore, we can precisely transfer the painting to the desired region while maintaining other parts unchanged.

The contributions of our paper can be summarized as follows.
\textbf{1)} We present a novel mesh-based neural implicit representation which aims to break the barrier between volumetric neural rendering and mesh-based 3D modeling and texturing workflow, and delivers a set of editing functionalities, including mesh-guided geometry editing, designated texture editing with texture swapping, filling and painting operations.
To make the representation locally editable both on geometry and texture, we design to encode the implicit field into mesh vertices, where each vertex possesses disentangled geometry and texture features of its local space.
\textbf{2)} We analyze the technical challenges and develop several techniques to enhance the spatial distinguishability with learnable sign indicators, ensure a steady training with distillation and fine-tuning mechanism, and improve the texture editing precision with spatial-aware optimization strategy.
\textbf{3)} Extensive experiments and impressive editing examples on both real and synthetic datasets demonstrate that our method achieves photo-realistic rendering quality, and is flexible and powerful at geometry and texture editing of the neural implicit field.

\section{Related Works}

\noindent\textbf{Mesh-based representation and rendering.}
In computer vision and graphics, polygon mesh has been widely used in 3D scene modeling and rendering~\cite{izadi2011kinectfusion,akenine2019real,liu2019soft}.
Traditional methods utilize multi-view geometry and numerical theories to reconstruct surface meshes of a captured scene~\cite{colmap,XuT19,kazhdan2006poisson,waechter2014let}.
Recently, more attention has been paid to neural network based scene reconstruction~\cite{murez2020atlas,sun2021neuralrecon} and texture learning~\cite{thies2019deferred,gao2020deferred,xiang2021neutex}.
However, existing mesh-based rendering pipelines usually require UV-mapping to build correspondences between meshes vertices and texture maps, which limits the applicability from representing scenes with complex topology and delicate structure.
Another line of methods uses MVS based mesh as a geometry proxy for image feature aggregation~\cite{riegler2020free,riegler2021stable}, but requires nearby source images to be warped back to the mesh surface and is not feasible for high-level editing operations.
Instead of storing textures in a flat 2D map or warping-based view synthesis, our method directly encodes appearance information on 3D vertices, and is more flexible in representing complex objects whose UV-maps are difficult to be unwrapped.

\noindent\textbf{Neural rendering.}
Given a set of image captures, neural rendering methods~\cite{dellaert2020neural,IDR} aim to render photo-realistic images of novel views.
NeRF~\cite{nerf} takes advantages of volume rendering to boost rendering quality, which inspires a lot of works, including surface reconstruction~\cite{unisurf,neus,volsdf}, human modeling~\cite{neural_actor,peng2021neural}, pose estimation~\cite{yen2021inerf},
scene understanding~\cite{yang2022_nr_in_a_room} and relighting~\cite{srinivasan2021nerv,zhang2021nerfactor,boss2021nerd,neural_outdoor_rerender}, \etc{}.
To further increase network capacity and reduce computation, many works propose to decompose scene into local representations, such as multiple tiny networks~\cite{reiser2021kilonerf}, point clouds~\cite{ost2021neural} and voxels~\cite{nsvf,yu2021plenoxels}.
Although these works explicitly encode scenes in a 3D spatial structure, they are not designed to be easily manipulated as polygon meshes, thus not capable of high-level applications like geometry and texture editing.

\noindent\textbf{Neural scene editing.}
Scene editing is a popular topic in computer vision and photography.
Early methods mainly focus on editing a single static view by inserting~\cite{karsch2011rendering}, compositing~\cite{perez2003poisson}, moving~\cite{kholgade20143d,shetty2018adversarial} objects or changing lighting~\cite{luo2020niid} for an existing photograph.
% Recently, 
With the development of neural rendering, many works start to edit scenes with movable~\cite{zhang2021editable,yang2021learning,guo2020object} and deformable~\cite{nerf_editing} objects, changeable colors, shapes~\cite{xie2021fig,liu2021editing} and textures~\cite{xiang2021neutex}.
However, existing methods are either limited to object-level rigid transformation~\cite{zhang2021editable,yang2021learning,guo2020object}, not generalize to out-of-distribution categories~\cite{deng2021deformed,liu2021editing,oechsle2019texture,sun2022fenerf,mvsnerf,wang2016unsupervised}, restricts its representation to simple shapes~\cite{xiang2021neutex} or orthographic projection~\cite{rematas2020neural}, or does not support fine-grained texture editing~\cite{liu2021editing,IDR,zhang2021nerfactor,srinivasan2021nerv,niemeyer2021giraffe,nerf_editing}.
By contrast, we pick up triangle mesh as a scaffold to encode the scene, since the mesh can be edited conveniently and intuitively in mature industry software, and the region of interest on the mesh can be precisely selected by vertices.
Built upon this, our method delivers the capability of non-rigid geometry editing and fine-grained texture editing.

\begin{figure*}[t!]
    \centering
    \includegraphics[width=1.0\linewidth, trim={0 0 0 0}, clip]{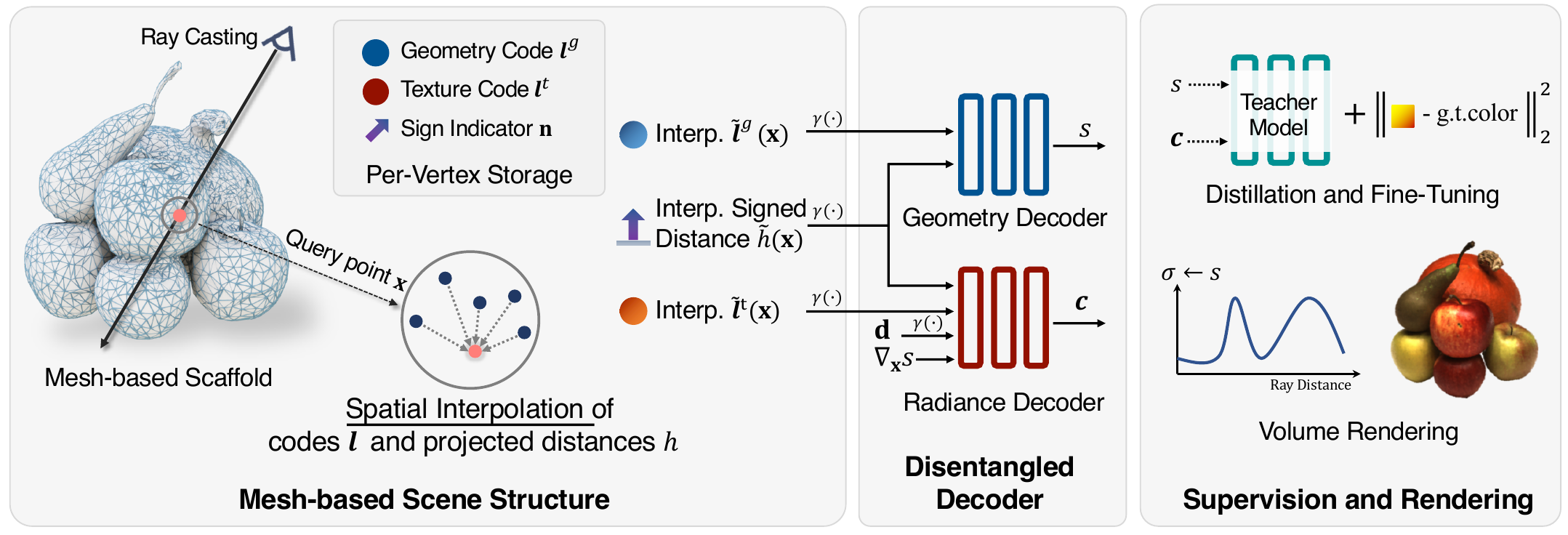}
    \caption{
    \textbf{Overview.} 
    We encode neural implicit field on a mesh-based scaffold, where each vertex possesses a geometry and texture code ${\bm{l}}^{g}, {\bm{l}}^{t}$, and a sign indicator $\textbf{n}$ for computing projected distance $h$.
    For a query point $\bm{x}$ along a casted camera ray, we retrieve interpolated codes and signed distances from the nearby mesh vertices, and forward to the geometry/radiance decoder to obtain SDF value $s$ and color $\textbf{c}$.
    }
    \label{fig:framework}
\end{figure*}

\section{Method}

We introduce NeuMesh, a novel scene representation that encodes neural implicit field at a mesh-based scaffold.
As demonstrated in Fig.~\ref{fig:framework}, instead of learning the entire scene as a whole in a coordinate-based network, we leverage 3D mesh structure by decomposing the scene into a set of local-vertex-bounded implicit fields (Sec.~\ref{ssec:repr}), where each vertex stores geometry and texture information of its neighboring local space.
Motivated by previous works~\cite{nerf,neus,reiser2021kilonerf}, we adopt the volume rendering technique to render pixels, and employ a distillation and fine-tuning training scheme to encode the neural implicit field into the mesh surface (Sec.~\ref{ssec:learn}).
During the rendering stage, we retrieve interpolated codes and learnable signed distances 
(\ie, projected distances to the mesh vertices, which complements spatial distinguishability) from the mesh, and use two separated MLPs to decode geometry (\ie, SDF values) and radiance color.
In this way, the scene representation is locally aligned to the mesh, and the geometry and color are encoded in two separated latent spaces, which naturally derives the approaches of mesh-guided geometry deforming and designatable texture editing (Sec.~\ref{ssec:edit}).

\subsection{Neural Mesh-based Implicit Field}
\label{ssec:repr}

\noindent\textbf{Mesh-based representation.}
As illustrated in Fig.~\ref{fig:framework}, we use a mesh-based scaffold to model the neural implicit field.
First, we reconstruct the target object using out-of-box NeuS~\cite{neus} and marching cubes~\cite{lorensen1987marching}, which yields a triangle mesh with about 50K$\sim$150K vertices.
Then, for each vertex $\textbf{v}$ on the mesh, we store a set of learnable parameters, including a geometry code $\bm{l}^{g}$, a texture code $\bm{l}^{t}$ and a sign indicator $\textbf{n}$ ($\textbf{n}$ helps to identify relative position, and will be introduced later).
In a typically volume rendering process that sample points $\textbf{x}$ along the ray, we first find $K$ nearest vertices $\{\textbf{v}_k | k = 1,2,...,K\}$ for each point $\textbf{x}$, and perform spatial interpolation to obtain the interpolated codes $\tilde{\bm{l}}^{g}(\textbf{x})$, $\tilde{\bm{l}}^{t}(\textbf{x})$ and signed distances $\tilde{h}(\textbf{x})$.
Specifically, we adopt inverse distance weighting based interpolation~\cite{qi2017pointnet}, as:
\begin{equation}
\label{eq:weight}
    \tilde{\bm{l}}(\textbf{x})=\frac{\sum^K_{k=1}w_k \bm{l}_k}{\sum^K_{k=1}w_k}, \;\; w_k=\frac{1}{||\textbf{v}_k-\textbf{x}||}.
\end{equation}
Then, we forward all these variables to geometry decoder $F_{G}$ and radiance decoder $F_{R}$ to obtain the SDF value $s = F_{G}\left(\tilde{\bm{l}}^g, \tilde{h}\right)$ and color $\bm{c} =F_{R}\left(\tilde{\bm{l}}^t,\tilde{h},\textbf{d},\nabla_\textbf{x} s\right)$ at point $\textbf{x}$,
where $\textbf{d}$ is the viewing direction, $\nabla_\textbf{x} s$ is the gradient of the SDF w.r.t query position.
Different from the previous methods~\cite{IDR,neus,volsdf},
we replace the global coordinate $\textbf{x}$ with locally retrieved codes $\tilde{\bm{l}}^g, \tilde{\bm{l}}^t$ and sign distances $\tilde{h}$, where $\tilde{h}$ complements spatial distinguishability without hurting the locality of the representation.
Note that we also apply positional encoding $\gamma(\cdot)$~\cite{nerf} to the interpolated codes, distance and direction before feeding them into the MLP, but we omit it in the equations for brevity.
Following the formulation of NeuS~\cite{neus} and quadrature rules~\cite{nerf}, we render the pixel $\hat{C}(\bm{r})$ with points $\{\mathbf{x}_i|i=1,...,N\}$ along the ray $\bm{r}$ as:
\begin{equation}
    \hat{C}(\bm{r}) = \sum_{i=1}^{N}    T_i \alpha_i {\mathbf{c}}_i, \; T_i = \prod_{j=1}^{i-1}(1-{\alpha}_j), \;
    {\alpha}_j = \max \left (\frac{\Phi_s(s_{i}) - \Phi_s(s_{i+1})}{\Phi_s(s_i)}, 0 \right),
\end{equation}
where $T$ is accumulated transmittance, $\Phi_s$ is the cumulative distribution of logistic distribution, and $\alpha$ is opacity derived from adjacent SDF.

\noindent\textbf{Learnable sign indicator for interpolated signed distance.}
To complement the spatial distinguishability of the network query along the direction perpendicular to the surface (\ie, inside or outside the mesh),
we introduce a \textit{learnable} sign indicator $\textbf{n}_k$ for each vertex $\textbf{v}_k$ that aids at computing interpolated signed distances for spatial query points.
Indeed, the sign indicator plays a similar role as vertex normal (\ie, initialized with vertex normal), but is continuously adjusted during the training process to best fit the target loss.
The computation of interpolated signed distance $\tilde{h}(\textbf{x})$ is defined as:
\begin{equation}
\begin{split}
    \tilde{h}(\textbf{x})=\frac{\sum^K_{k=1}w_k h_k}{\sum^K_{k=1}w_k}, \; h_k = \textbf{p}_k \cdot \frac{\omega^n \textbf{n}_k+ \omega^p_k\textbf{p}_k}{\omega^n + \omega^p_k}, \;\textbf{p}_k=\textbf{x} - \textbf{v}_k,
\end{split}
\end{equation}
where $w_k$ is inverse distance weighting as defined in Eq.~\eqref{eq:weight},
$\omega^n$ and $\omega^p_k$ controls the influence between sign indicator and point-to-vertex vector $\textbf{p}_k$, and we empirically set $\omega^n=0.1, \omega^p_k=||\textbf{p}_k||$.
Intuitively, when the sample points are far from the surface, $\tilde{h}(\textbf{x})$ is numerically close to the point-to-surface distance; otherwise, when the sample points are getting close to the surface, $\tilde{h}(\textbf{x})$ would be gradually perturbed by learnable sign indicators.

\subsection{Optimizing Mesh-based Implicit Field}
\label{ssec:learn}

\noindent\textbf{Distillation and fine-tuning.}
We observe that training NeuMesh from scratch leads to artifacts and converges to sub-optimal results (see Fig.~\ref{fig:ablation_vis} and Tab.~\ref{tab:ablation}).
Inspired by Reiser \etal~\cite{reiser2021kilonerf}, we apply a distillation and fine-tuning training scheme, \ie, we supervise NeuMesh simultaneously with the output from a coordinate-based teacher model (\eg, NeuS), and also the images.
For a batched training rays $\bm{r}\in R$, we defined the distillation loss $L_d$ and photometric fine-tuning loss $L_f$ as:
\begin{equation}
    \mathcal{L}_{\text{d}} = \sum_{\bm{r}\in R}\sum_{i\in N}||{s}_i - s^t_i|| + ||{\bm{c}}_i-\bm{c}^t_i|| ,\;\; 
    \mathcal{L}_{\text{f}} =\sum_{\bm{r}\in R}||\hat{C}(\bm{r})-C(\bm{r})||^2_2
\end{equation}
where ${s}_i^t$ and $\bm{c}_i^t$ are the SDF value and color from the teacher model, and $C(\bm{r})$ is the ground-truth pixel color from images.
By leveraging distillation and fine-tuning, we smoothly transfer a pure MLP-based neural implicit model into a flexible and editable mesh-based representation, and the final model even produces better appearance details, as shown in our experiment (Sec.~\ref{ssec:compare_quality}).

\noindent\textbf{Regularization.}
As introduced in Sec.~\ref{ssec:repr}, we dynamically adjust a set of per-vertex sign indicators during the training process.
To ensure a smooth convergence, we empirically apply a regularization to the sign indicator by slightly encouraging them being close to pre-computed vertex normal $\bm{n}^t$, as: $\mathcal{L}_{\text{rs}}= \sum_{k} ||\bm{n}_k-\bm{n}^t_k||_2^2$.
Besides, as suggested by Gropp \etal~\cite{igr}, we add an Eikonal loss to regularize the norm of the spatial gradients to 1, as: $\mathcal{L}_{\text{re}}=\sum _{k}||\left\lVert\nabla_{\textbf{x}_k} s_k\right\rVert-1||^2_2$.

The final loss is then defined as:
\begin{equation}
    \mathcal{L}_{\text{total}} = 
    \lambda_{\text{d}} \mathcal{L}_{\text{d}} 
    + \lambda_{\text{f}} \mathcal{L}_{\text{f}} 
    + \lambda_{\text{rs}} \mathcal{L}_{\text{rs}} 
    + \lambda_{\text{re}} \mathcal{L}_{\text{re}},
\end{equation}
where we set $\lambda_{\text{d}}=1.0$, $\lambda_{\text{f}}=1.0$, $\lambda_{\text{rs}}=0.01$ and $\lambda_{\text{re}}=0.01$.

\subsection{Mesh-guided Geometry Editing}
\label{ssec:edit}

\begin{figure}[t!]
    \centering
    \includegraphics[width=1.0\linewidth, trim={0 0 0 0}, clip]{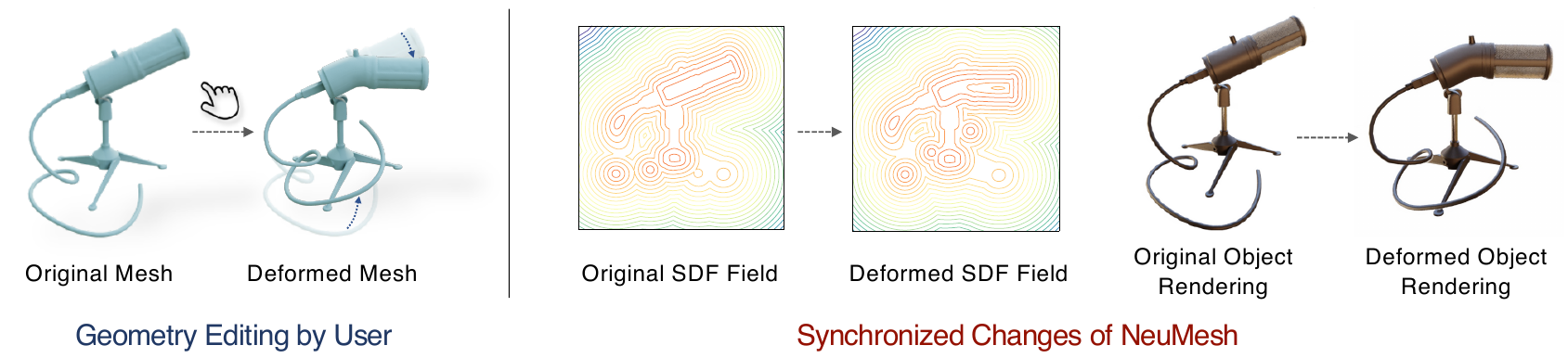}
    \caption{
    \textbf{Mesh-guided Geometry editing.}
    By simply deforming the corresponding mesh, the change will synchronously take effect on the implicit field, and the rendered object will also be deformed accordingly.
    }
    \label{fig:geo_edit_process}
\end{figure}

% \noindent\textbf{Mesh-guided geometry editing.}
In NeuMesh, since the neural implicit field has been tightly aligned to the mesh surface, any manipulation on mesh vertices would directly take effect on the field and the volume rendering results.
Therefore, to perform geometry editing with a NeuMesh-based scene, users are only required to edit the corresponding mesh, which can be easily accomplished by interactively moving a few vertices with out-of-box mesh deforming methods (\eg, as-rigid-as-possible, or ARAP~\cite{arap}), or 3D modeling software like Blender.
We show an example of the geometry editing in Fig.~\ref{fig:geo_edit_process}, where we first deform the microphone by bending its head and lifting the wire on the corresponding mesh.
Then, to maintain the local consistency of signed distance (Sec~\ref{ssec:repr}), for each transformed vertex, we also compute a relative rotation of the surface normal and compensate the rotation according to its sign indicator (Sec.~\ref{ssec:repr}).
Without any fine-tuning, the microphone's implicit field has been deformed in the meantime, and we can easily render the deformed view (see Fig.~\ref{fig:geo_edit_process}).
Please refer to the supplementary materials for more details.

\begin{figure}[!t]
    \centering
    \includegraphics[width=1.0\linewidth, trim={0 0 0 0}, clip]{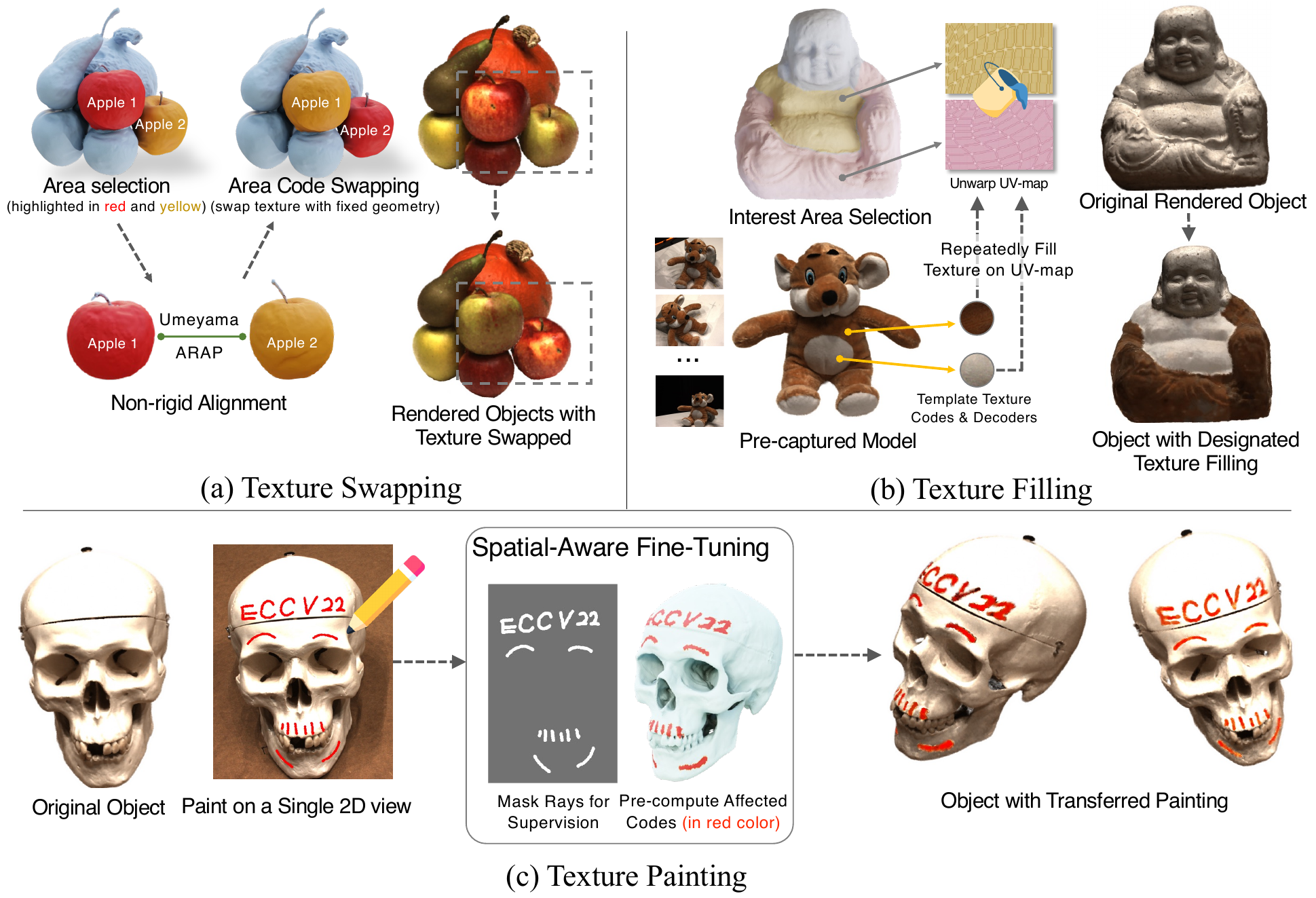}
    \caption[]{
    \textbf{Designatable Texture editing.} By exchanging texture codes (and decoders), our representation delivers various texture editing pipelines on neural implicit field.
    \footnotemark[1]
    }
    \label{fig:tex_edit_process}
\end{figure}

\footnotetext[1]{Icon credit: Flaticon~\cite{flaticon}}

\subsection{Designatable texture editing}
Until then, texture editing on the neural implicit model is still an open problem.
Previous methods tend to replace the entire materials by swapping the appearance branch~\cite{IDR,zhang2021nerfactor,srinivasan2021nerv}, changing a uniformed color~\cite{liu2021editing}, or learning an editable UV mapping for simple and plump shapes~\cite{xiang2021neutex}.
However, in real texturing of 3D modeling software, artists are used to working with a mesh-based workflow, which allows them to select a partial region of an object and modify it with arbitrary colors and material properties.
We propose to mimic such pipelines by introducing a designatable texture editing, where the selection of mesh vertices can be used to precisely guide the texture editing on the region of interest.
The core step of our texture editing is that we update the latent texture code $\bm{l}^t$ (`material properties') and the binding decoder $F_R$ (`rendering palette') at the selected region.
As shown in Fig.~\ref{fig:tex_edit_process}, we deliver three ways of texture editing:

\noindent\textbf{1) Texture swapping} by exchanging textures between two objects through 3D geometry (\eg, swapping textures of two apples in Fig.~\ref{fig:tex_edit_process}).
Users are first asked to mark out the source and target object on the mesh, which can be done by mature 3D model software, or point-based instance segmentation~\cite{wang2018sgpn}.
Then, given a putative point matches with interactive annotation~\cite{zhou2018open3d}, we perform non-rigid 3D alignment to the source and target object with Umeyama~\cite{umeyama} and ARAP~\cite{arap}.
Finally, we transfer texture codes by assigning each target vertex with code interpolated from nearby source vertices.

\noindent\textbf{2) Texture filling} by filling a targeting object area with repeated textures from a pre-captured model (\eg, assigning part of Buddha with two furry materials from a teddy bear as shown in Fig.~\ref{fig:tex_edit_process}).
In real applications, artists might want to try out some materials from a daily captured scene or pre-built material library, or want to fill some areas (\eg, floor and walls) with uniformed materials.
Therefore, we build a compatible workflow for the standard texturing operation, where we first construct a UV map for the user-interest areas, and 
then repeatedly fill the mapped vertices  (\eg, chest and cloth of Buddha) with template textures from a pre-captured model (\eg, gray and brown hairs from the teddy bear).

\noindent\textbf{3) Texture painting} from a single 2D view to the 3D field.
Users paint an arbitrary pattern or put some text on a captured image, and we can transfer these paintings into the 3D neural implicit field and freely preview in rendered novel views (\eg, painting ECCV logo on a skull in Fig.~\ref{fig:tex_edit_process}).
Compared to NeuTex~\cite{xiang2021neutex} that might be difficult to edit on the desired position due to distorted UV-mapping, our method delivers a more natural editing way, \ie, what you draw and see is what you get.
However, it is not trivial to precisely control the painting transferring with only one image, since the overfitting of a single image might lead to appearance drifting at unconstrained views, which inevitably introduces artifacts in rendered novel views.
To tackle this issue, we propose a \textit{spatial-aware optimization mechanism}.
Specifically, we first shoot rays through the painted pixels to obtain the surface points and find the affected texture codes of nearby vertices around the points.
During the fine-tuning stage, we optimize by minimizing photometric loss of rendered pixels and painted pixels, and only backward gradient of these codes while detaching the others.
Besides, to improve the training efficiency and the painting consistency across views, we restrict training rays inside a slightly dilated paint mask, and also augment with random viewing directions at the input to the radiance decoder.

Please refer to the supplementary materials for more details.

\section{Experiments}

\subsection{Datasets}
\label{ssec:dataset}
We evaluate our method on the real captured DTU~\cite{dtu} dataset and NeRF 360$^{\circ}$ Synthetic dataset.
For the DTU dataset, we follow the setting of IDR~\cite{IDR} by using 15 scenes with images of $1600\times 1200$ resolution and foreground masks for experiments.
To facilitate the metric evaluation for both rendering and mesh quality, we randomly select 10\% images as test split and use the remaining images for training.
For NeRF 360$^{\circ}$ Synthetic dataset, we follow the official split and choose 4 representative scenes for evaluation, including thin structures (Mic), complex shapes (Lego) and rich textures (Chair and Hotdog).

\begin{figure}[!t]
    \centering
    \includegraphics[width=1.0\linewidth, trim={0 0 0 0}, clip]{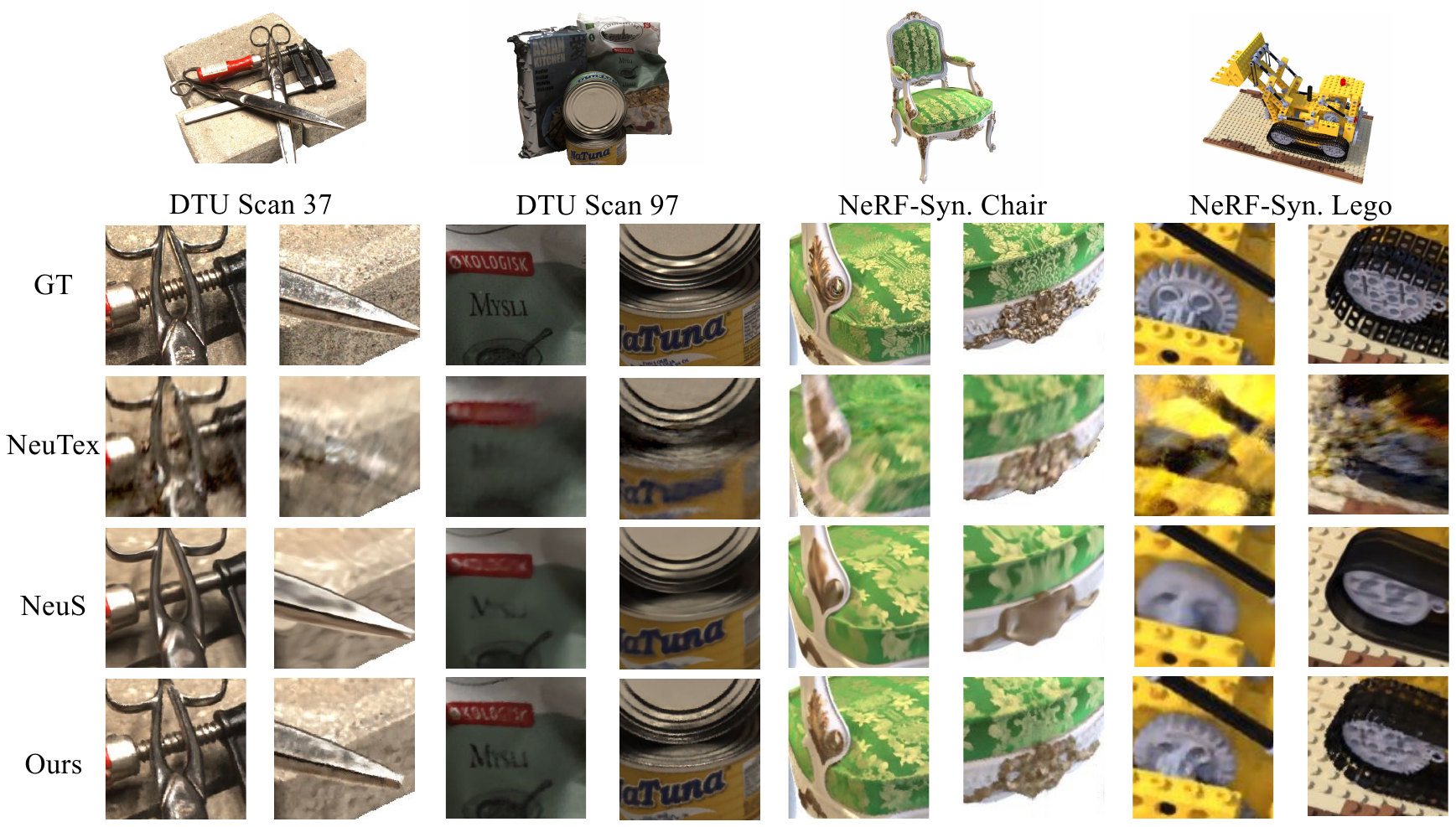}
    \caption{
    We show rendering examples of NeuTex~\cite{xiang2021neutex}, NeuS~\cite{neus} and our method on the DTU dataset and the NeRF 360$^\circ$ Synthetic dataset.
    }
    \label{fig:render_quality}
\end{figure}

\begin{table}[tb]
\centering
\resizebox{1.0\linewidth}{!}{
\tabcolsep 13pt
\begin{tabular}{lcccccc}
\toprule
\multicolumn{1}{c}{\multirow{2}{*}{Methods}} & \multicolumn{3}{c}{DTU} & \multicolumn{3}{c}{NeRF 360$^{\circ}$ Synthetic} \\ \cmidrule(lr){2-4} \cmidrule(lr){5-7}  
\multicolumn{1}{c}{} & \multicolumn{1}{l}{PSNR $\uparrow$} & \multicolumn{1}{l}{SSIM $\uparrow$} & \multicolumn{1}{l}{LPIPS $\downarrow$} & \multicolumn{1}{l}{PSNR $\uparrow$} & \multicolumn{1}{l}{SSIM $\uparrow$} & \multicolumn{1}{l}{LPIPS $\downarrow$} \\ \hline
NeuTex~\cite{xiang2021neutex} & 26.080 & 0.893 & 0.196 & 25.718 & 0.914 & 0.109 \\
NeuS~\cite{neus} & 26.352 & 0.909 & 0.176 & 30.588 & \textbf{0.960} & 0.058 \\
Ours & \textbf{28.289} & \textbf{0.921} & \textbf{0.117} & \textbf{30.945} & 0.951 & \textbf{0.043} \\
\bottomrule
\end{tabular}
}
\caption{
We compare rendering quality with NeuS~\cite{neus} and NeuTex~\cite{xiang2021neutex} on the DTU dataset and the NeRF 360$^{\circ}$ Synthetic dataset.
}
\label{tab:render}
\end{table}

\subsection{Comparison of Rendering and Mesh Quality}
\label{ssec:compare_quality}

We first compare the rendering and mesh reconstruction quality of our representation with the baseline method NeuS~\cite{neus} and the SOTA texture-editable implicit neural rendering method NeuTex~\cite{xiang2021neutex}.
Following previous works~\cite{nerf,neus,IDR}, we use PSNR, SSIM and LPIPS to measure the rendering quality, and use Chamfer distance to measure the reconstructed mesh quality.
Please note that for mesh quality comparison, we use a subset (training split) of images, while NeuS takes all images for training in their paper, so the result is slightly different.
As demonstrated in Fig.~\ref{fig:render_quality} and Tab.~\ref{tab:render}, our method is comparable or even better than NeuS and NeuTex on rendering quality.
To achieve texture editing, NeuTex attempts to encode all the textures in a single continuous UV space, which works for plump objects (\eg, plush toys or Buddah as shown in its paper) but struggles to reconstruct objects with complex shapes (\eg, scissors in
 DTU Scan 37 and gears in NeRF-Synthetic Lego as shown in Fig.~\ref{fig:render_quality}).
We consider that because NeuTex tries to memorize all textures in the single continuous UV-map by using a simplified Atlas-Net~\cite{groueix2018papier} (\ie, one atlas), which limits its representation of complex shapes.
For NeuS, as it pursues better mesh reconstruction quality than novel view synthesis, the details of rendered images are slightly blurred and smoothed.
By contrast, our representation not only delivers the capability of geometry and texture editing, but also shows clear appearance detail (see Fig.~\ref{fig:render_quality}) and maintains mesh quality on par with NeuS (see Tab.~\ref{tab:mesh}).

\begin{figure}[t!]
    \centering
    \includegraphics[width=1.0\linewidth, trim={0 0 0 0}, clip]{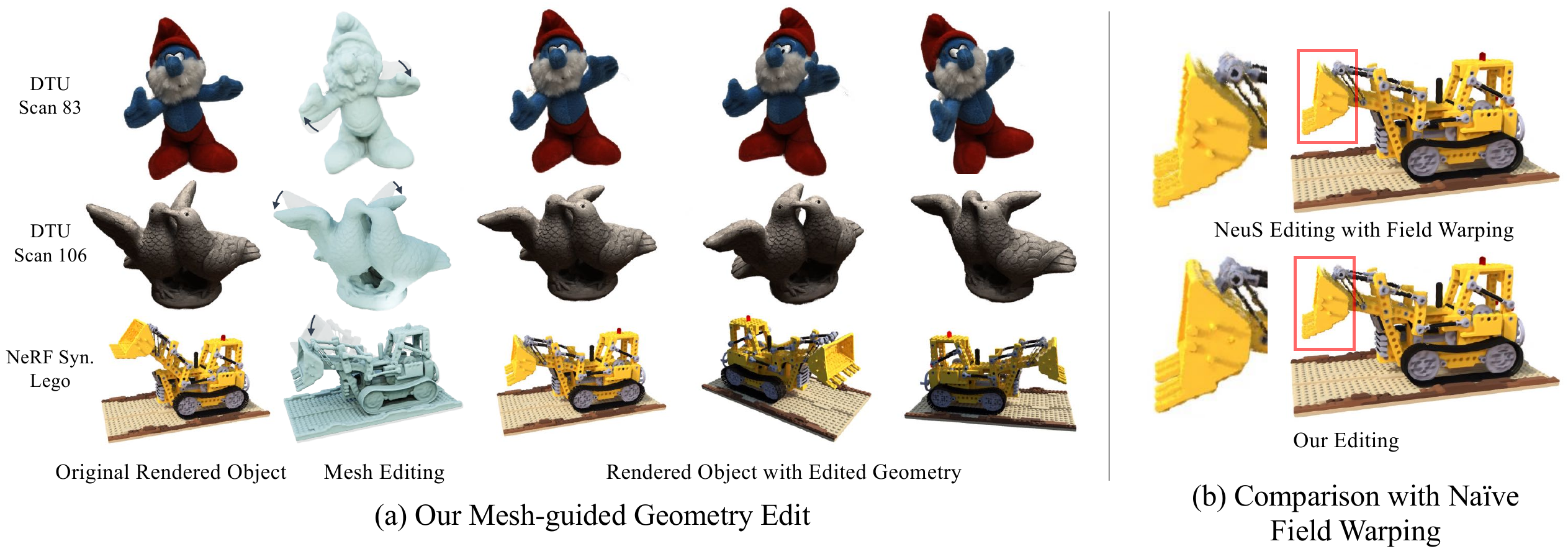}
    \caption{
    We show examples of mesh-guided geometry editing in (a) and also compare with na\"ive field warping solution in (b).
    }
    \label{fig:geo_edit}
\end{figure}

\begin{table}[tb]
\centering
\resizebox{1.0\linewidth}{!}{
\tabcolsep 3pt
\begin{tabular}{lcccccccccccccccc}
\toprule
 & \multicolumn{16}{c}{DTU Scan ID} \\ \midrule
Method & \multicolumn{1}{r}{24} & \multicolumn{1}{r}{37} & \multicolumn{1}{r}{40} & \multicolumn{1}{r}{55} & \multicolumn{1}{r}{63} & \multicolumn{1}{r}{65} & \multicolumn{1}{r}{69} & \multicolumn{1}{r}{83} & \multicolumn{1}{r}{97} & \multicolumn{1}{r}{105} & \multicolumn{1}{r}{106} & \multicolumn{1}{r}{110} & \multicolumn{1}{r}{114} & \multicolumn{1}{r}{118} & \multicolumn{1}{r}{122} & \multicolumn{1}{r}{Avg.} \\ \midrule
NeuTex~\cite{xiang2021neutex} & 2.078 & 5.038 & 3.477 & 1.039 & 3.744 & 2.078 & 3.201 & 2.163 & 5.104 & 1.828 & 1.951 & 4.319 & 1.177 & 3.100 & 1.921 & 2.815  \\
NeuS*~\cite{neus} & 1.544 & \textbf{1.224} & 1.065 & \textbf{0.665} & 1.286 & \textbf{0.825} & 0.904 & 1.350 & 1.320 & 0.855 & 0.987 & 1.328 & \textbf{0.487} & \textbf{0.636} & \textbf{0.678} & 1.010 \\
Ours & \textbf{1.112} & 1.262 & \textbf{0.988} & 0.674 & \textbf{1.224} & 0.835 & \textbf{0.878} & \textbf{1.232} & \textbf{1.304} & \textbf{0.741} & \textbf{0.963} & \textbf{1.239} & 0.558 & 0.645 & 0.739 & \textbf{0.960}  \\ \bottomrule
\end{tabular}
}
\caption{
We compare mesh quality (Chamfer distance) with NeuS~\cite{neus} on the DTU dataset.
Note that we use training split of images instead of full images, so the result of NeuS~\cite{neus} is different from the original paper.
}
\label{tab:mesh}
\end{table}

\subsection{Experiment on Geometry Editing}
\label{ssec:expr_geo_edit}

We now show the result of mesh-guided geometry editing in Fig.~\ref{fig:geo_edit} (a), where we simply deform meshes with Blender, and the rendered objects are deformed simultaneously.
Since an implicit field can be trivially deformed with non-rigid warping, we also compare our editing with a na\"ive field warping solution that is directly applied to NeuS, which bends the query points from the deformed space to the original space by computing interpolated warping with the offsets from 3 nearest vertices of the extracted mesh.
As shown in Fig.~\ref{fig:geo_edit} (b), the object boundary of the field warping results is much jaggier than ours, which proves the necessity of our mesh-based representation on this task.

\begin{figure}[!htbp]
    \centering
    \includegraphics[width=1.0\linewidth, trim={0 0 0 0}, clip]{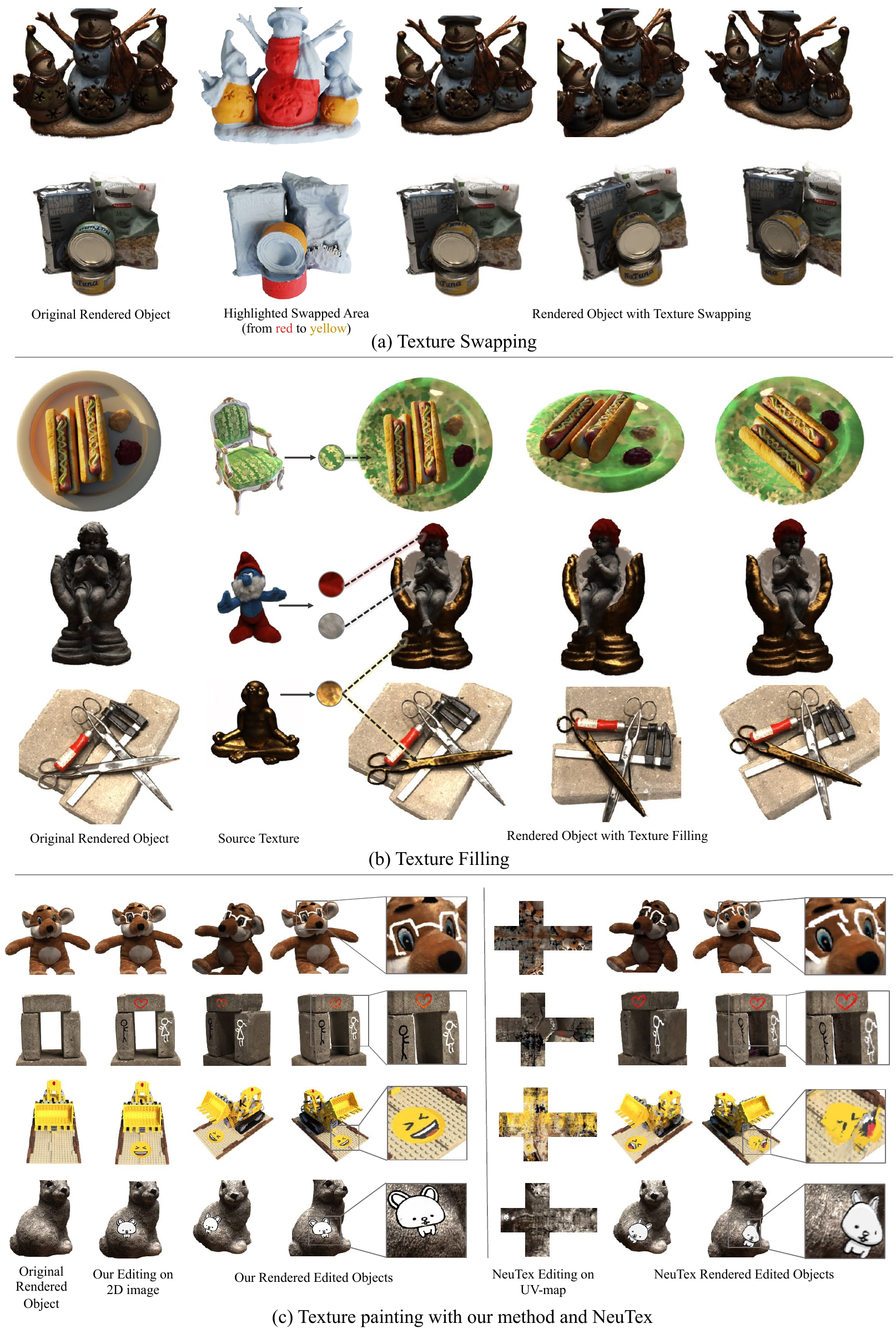}
    \caption{
    We show texture editing examples on the DTU dataset and the NeRF 360$^\circ$ Synthetic dataset.
    }
    \label{fig:tex_edit}
\end{figure}

\subsection{Experiment on Texture Editing}
\label{ssec:expr_tex_edit}

To the best of our knowledge, prior to our work, only NeuTex~\cite{xiang2021neutex} supports texture editing of the neural implicit field by painting on 2D UV texture.
However, due to the distorted UV mapping, we find NeuTex hard to perform all the editing operations like ours, so we only compare it on the texture painting task.

\noindent\textbf{Texture swapping.}
We present 2 examples of texture swapping in Fig.~\ref{fig:tex_edit} (a), where the textures of the snowman's body and the packaging of cans have been seamlessly swapped, and even the details (\eg, texts on the cans) have been clearly transferred into the target object, while the geometry is kept unchanged.
This demonstrates that our representation successfully disentangles geometry and texture in two spaces, and the disentangled texture representation can be seamlessly integrated into new shapes.

\noindent\textbf{Texture filling.}
We show 3 examples of texture filling in Fig.~\ref{fig:tex_edit} (b), in which the targeting areas are repeatedly filled with template texture code and decoder from previously captured source models.
It is worth noting that even though the source template only covers a small area of texture codes, we can still observe view-dependent effects (\eg, golden metal naturally exhibits specular reflections at different views).

\noindent\textbf{Texture painting.}
In Fig.~\ref{fig:tex_edit} (c), we exhibit 4 examples of texture painting, and also conduct similar editing with NeuTex~\cite{xiang2021neutex} by painting on the unwrapped UV-map.
For NeuTex, as the learned texture mapping is somehow irregular and distorted (\eg, the head of the teddy bear is separated in UV-map), we find it hard to paint at the desired location.
In the second row (painting on bricks), we have to adjust the painting position back and forth to get a reasonable editing result.
Besides, due to the mapping issue explained in Sec.~\ref{ssec:compare_quality}, NeuTex cannot picture a clear result when editing on complex shapes (Lego in the second row).
On the contrary, our method offers a user-friendly editing pipeline by directly painting on 2D images and then transferring the painting into the 3D implicit field.

\subsection{Ablation Studies}
\label{ssec:ablation}

\begin{figure}[t!]
    \centering
    \includegraphics[width=1.0\linewidth, trim={0 0 0 0}, clip]{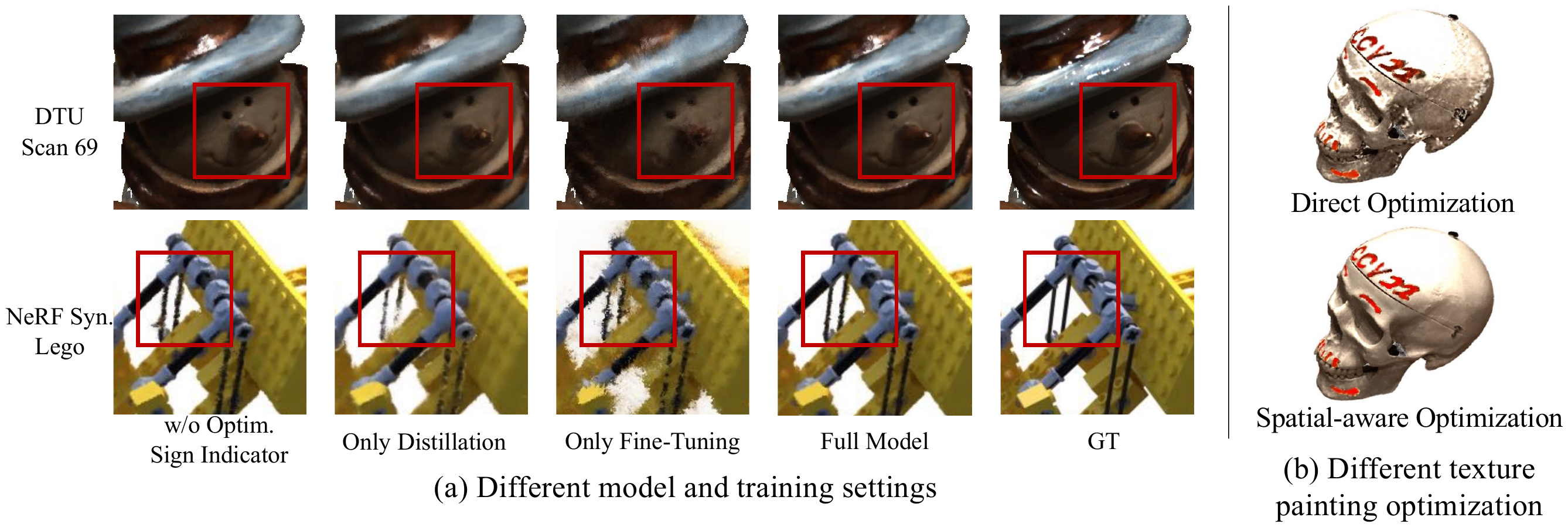}
    \caption{
    We show the rendering results with different settings in (a) and also show the effectiveness of spatial-aware optimization for texture painting in (b).
    }
    \label{fig:ablation_vis}
\end{figure}

\begin{table}[tb]
\centering
\resizebox{1.0\linewidth}{!}{
\tabcolsep 10pt
\begin{tabular}{lcccccc}
\toprule
\multicolumn{1}{c}{\multirow{2}{*}{Config.}} & \multicolumn{3}{c}{DTU 69} & \multicolumn{3}{c}{NeRF 360$^{\circ}$ Synthetic Lego} \\ \cmidrule(lr){2-4} \cmidrule(lr){5-7}  
\multicolumn{1}{c}{} & \multicolumn{1}{l}{PSNR $\uparrow$} & \multicolumn{1}{l}{SSIM $\uparrow$} & \multicolumn{1}{l}{LPIPS $\downarrow$} & \multicolumn{1}{l}{PSNR $\uparrow$} & \multicolumn{1}{l}{SSIM $\uparrow$} & \multicolumn{1}{l}{LPIPS $\downarrow$} \\ \hline
w/o Optim. Sign Indicator & 26.633 & 0.940 & 0.119 & 27.631 & 0.922 & 0.053 \\
Only Distillation & 26.599& 0.936 & 0.144 & 25.606 & 0.901 & 0.081 \\
Only Fine-Tuning & 26.258 & 0.926 & 0.132 & 23.583 & 0.876 & 0.135 \\
Full Model & \textbf{27.254} & \textbf{0.946} & \textbf{0.113} & \textbf{27.881} & \textbf{0.926} & \textbf{0.046} \\
\bottomrule
\end{tabular}
}
\caption{
We perform ablation studies on the model design and training strategy with DTU Scan 69 and NeRF 360$^\circ$ Synthetic Lego.
}
\label{tab:ablation}
\end{table}

\noindent\textbf{Learnable sign indicator.}
We first inspect the effectiveness of the proposed learnable sign indicator in each vertex.
Specifically, we set sign indicators as constant vertex normal without adjusting during the training process and evaluate the model both qualitatively and quantitatively.
As demonstrated in Fig.~\ref{fig:ablation_vis} (a) and Tab.~\ref{tab:ablation} (first row), online adjusting sign indicators consistently improves the image quality.
By the way, we notice that the PSNR improvement on real data (DTU Scan 69) is more significant than the synthetic one (Lego). We consider that the mesh quality (and vertex normal) of real data is worse than the synthetic data due to sensor noises, which degrades the rendering quality, while the learnable sign indicator helps to mitigate this issue.

\noindent\textbf{Distillation and fine-tuning training scheme.}
We then study the necessity of distillation and fine-tuning training scheme by ablating one of them during model training.
As shown in Fig.~\ref{fig:ablation_vis} (a) and Tab.~\ref{tab:ablation} (second and third row),
by enabling distillation only, the rendered image is blurry than the full model ones.
When using fine-tuning without distillation, the rendering result ends up with noticeable artifacts.
These results suggest that both distillation and fine-tuning are indispensable when training our mesh-based representation.

\noindent\textbf{Spatial-aware optimization in texture painting.}
We also evaluate the proposed spatial-aware optimization in the texture painting task and visualize the comparison in Fig.~\ref{fig:ablation_vis} (b).
It is clear that when na\"ively optimizing painting with a single image, the model will overfit to the specific viewpoint, and the change to the texture codes might break the appearance consistency, which results in visual artifacts when rendering the implicit field from the side view.
By introducing a spatial-aware optimization mechanism, we successfully avoid such artifacts and obtain the modified field while maintaining other parts untouched.

\section{Conclusion}

We have proposed a novel mesh-based neural representation, which supports high-fidelity volume rendering, and flexible geometry and texture editing.
Specifically, we encode the neural implicit field into a mesh scaffold, where each mesh vertex possesses learnable geometry and texture code for its neighboring local space.
One limitation of our method is that we do not model fine-grained lighting effects such as shadowing and specular reflection of a certain lighting environment, which can be improved by introducing material and lighting estimation in future works.
Besides, due to the reliance on mesh scaffold, we cannot represent objects that fails during reconstruction (\eg, smoke or liquid).

% ~\\
\vspace*{\baselineskip}

\noindent\textbf{Acknowledgment.} This work was partially supported by NSF of China~(No. 61932003, No. 62102356).

\clearpage
% ---- Bibliography ----
%
% BibTeX users should specify bibliography style 'splncs04'.
% References will then be sorted and formatted in the correct style.
%
\bibliographystyle{splncs04}
\bibliography{main}

\clearpage

\appendix

\renewcommand\thesection{\Alph{section}}
\renewcommand\thetable{\Alph{table}}
\renewcommand\thefigure{\Alph{figure}}

{\noindent \Large \bf Supplementary Material\par}

\vspace{0.5em}

\renewcommand\thesection{\Alph{section}}
\renewcommand\thetable{\Alph{table}}
\renewcommand\thefigure{\Alph{figure}}

In this supplementary material, we describe more details of our method, including model architecture in Sec.~\ref{sec:model_arch}, geometry editing in Sec.~\ref{ssec:detail_geo_edit}, and texture editing in Sec.~\ref{ssec:detail_tex_edit}.
Besides, we also provide more discussions including limitations in Sec.~\ref{sec:supp_discussion} and experiment results in Sec.~\ref{sec:expr}.

\section{Model Architecture}
\label{sec:model_arch}

\begin{figure}[h]
    \centering
    \includegraphics[width=1.0\linewidth, trim={0 0 0 0}, clip]{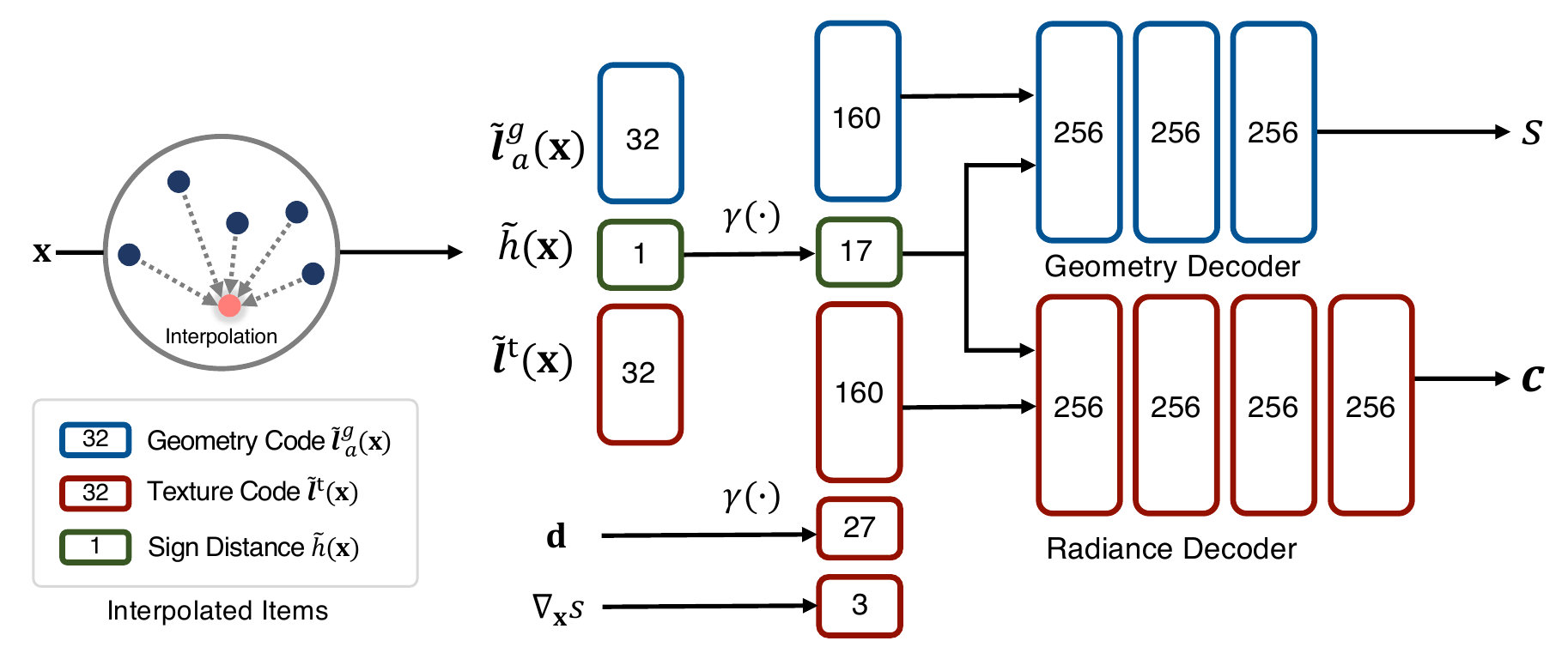}
    \caption{
    \textbf{The model architecture of NeuMesh.}
    }
    \label{fig:supp_architecture}
\end{figure}

The detailed model architecture is shown in Fig.\ref{fig:supp_architecture}.
To begin with, we first extract a triangle mesh with marching cubes~\cite{lorensen1987marching} from NeuS's~\cite{neus} SDF field, where we set the voxel resolution as 256 and the spatial range as $[-1, 1]$.
Then, for each query point $\mathbf{x}$, we find $K$ nearest vertices (with $K=8$ in our experiments) and obtain the interpolated geometry code (32 dimensions), texture code (32 dimensions) and learnable signed distance (scalar) from these vertices.
Before feeding into the network, we apply positional encoding to the signed distances (with 8 frequencies), interpolated codes (with 2 frequencies) and viewing directions (with 4 frequencies).
The geometry decoder and the radiance decoder are constructed with a MLP of 3 / 4 hidden layers and 256 hidden sizes, and we use SoftPlus / ReLU activation, respectively.
During the rendering stage, we first sample 64 coarse points along the ray and adopt a progressive up-sampling strategy from Wang \etal ~\cite{neus} to guide the sampling of 64 fine points, which yields 128 samples for each ray.
Besides, to accelerate rendering and training, we pre-compute near-far bound for each ray by counting the minimum and maximum distances of ray-to-mesh intersections.

\section{Implementation Details}
\label{sec:impl}

\subsection{Training Details}
\label{ssec:train}

As introduced in our main paper, we adopt a distillation and fine-tuning training scheme.
Practically, for each object, we first train a teacher model (\ie, NeuS~\cite{neus}).
Then, we optimize codes and decoders with output from the teacher model and the images.
During the training process, we use a batch size of 512 rays on a single Nvidia RTX3090-24G GPU, where the queried color and SDF value for each sample point will also be supervised with the output from the teacher model (a.k.a distillation loss in Sec.~3.2 Eq.(4)).
We adopt the Adam optimizer with an initial learning rate of 0.0005 and a cosine annealing scheduler with 5000 warm-up steps.
The training process takes about 16 hours for each model.
Besides, to train on the DTU dataset that contains unbounded background, we follow previous works~\cite{IDR,neus} by taking foreground masks into the supervision with a binary cross-entropy loss.

\begin{figure}[t!]
    \centering
    \includegraphics[width=1.0\linewidth, trim={0 0 0 0}, clip]{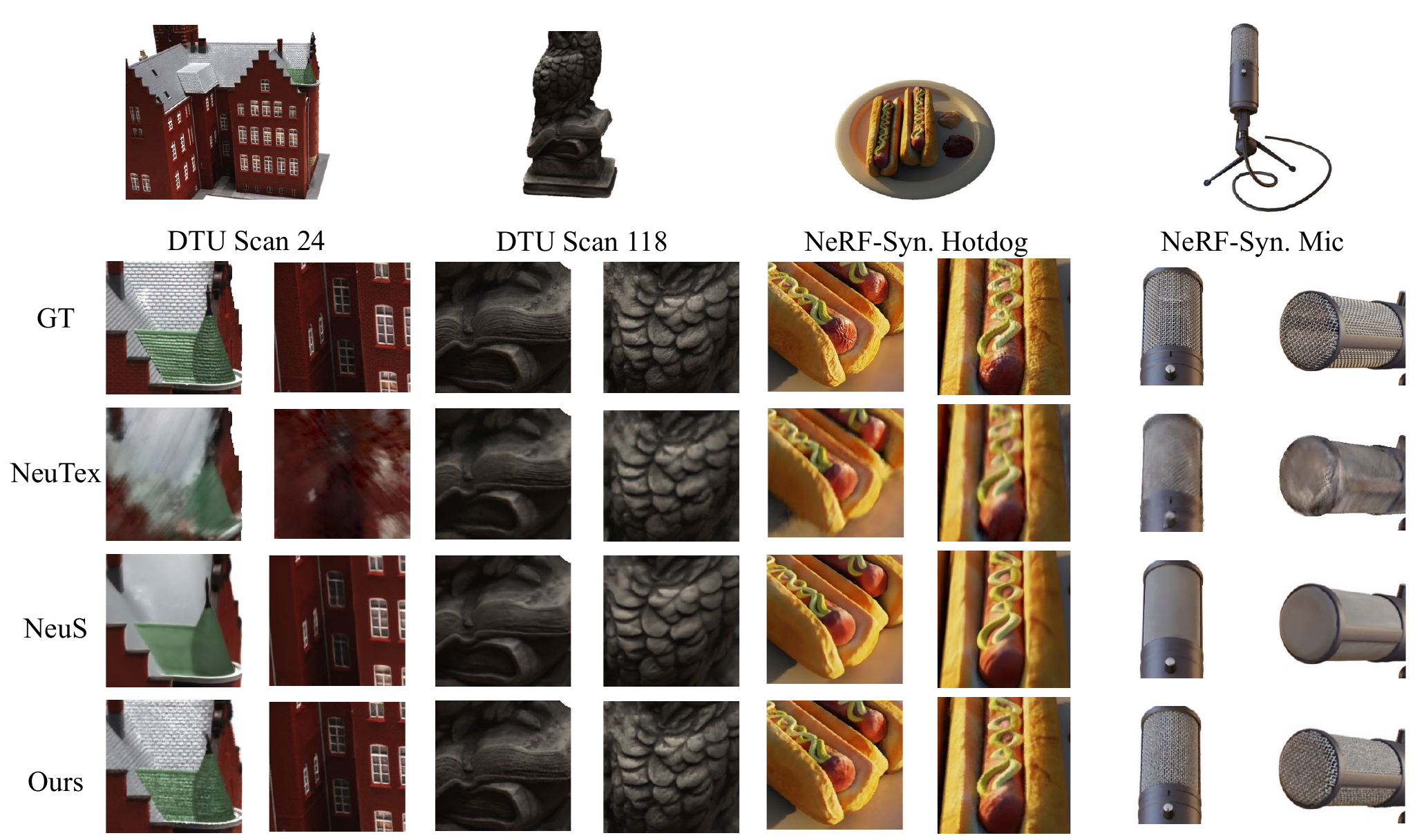}
    \caption{
    \textbf{
    We show more comparison of rendering quality on the DTU dataset and the NeRF 360$^\circ$ Synthetic dataset. 
    } Our rendering results show better appearance details than NeuS and NeuTex (\eg, the roof at DTU Scan 37, and the metal grids at NeRF-Synthetic Mic).
    }
    \label{fig:supp_render_compare}
\end{figure}
\subsection{Details of Geometry Editing}
\label{ssec:detail_geo_edit}

With our mesh-based representation, deforming a neural implicit field is as simple as deforming the corresponding mesh scaffold. 
The only thing to note is to keep the local consistency of the learnable signed distances (Sec.~3.1), \ie, the interpolated signed distance of the locally deformed or rotated region should keep the same.
To achieve this goal, we simply compensate the rotation of the surface normal to the learnable signed indicator $\tilde{\bm{h}}(\textbf{x})$, as: 
$\tilde{\bm{h}}'(\textbf{x})=\tilde{\bm{h}}(\textbf{x}) + \delta h_x$,
where $\delta h_x$ is the relative difference of vertex normal (averaged from the nearby surface normal) from the original mesh to the deformed mesh, and $\tilde{\bm{h}}'(\textbf{x})$ is the compensated signed indicator.

\subsection{Details of Texture Editing}
\label{ssec:detail_tex_edit}

Since our representation disentangles textures into locally bounded texture codes saved on mesh vertices, texture editing for a neural implicit field can be accomplished by updating or optimizing texture codes (and the binding encoders) for the region of interest.

\noindent\textbf{Texture swapping.}
We can easily swap textures of two areas by swapping texture codes on the surface, as long as we find the correspondence from the source area's vertices to the target area's vertices.
To this end, we provide a solution to perform texture swapping on two areas that can be reasonably aligned but with slightly different shapes (\eg, two apples in Fig.4 (a) from the main paper).
In practice, we first choose source and target areas by selecting mesh vertices with Blender, and annotate 4$\sim$9 point correspondences with our scripts between these two areas.
Note that this can also be automated with point cloud or image segmentation tools when deploying to user-friendly applications.
Then, we perform non-rigid mesh alignment by solving scaled transformation with Umeyama~\cite{umeyama} between point correspondences, and then feed the point residual to ARAP~\cite{arap}, so as to deform the source area to the target area.
Finally, we update the texture codes on the target area by assigning interpolated code (with inverse distance weighting) from 4 nearest deformed source vertices.

\noindent\textbf{Texture filling.}
By leveraging NeuMesh, our model supports filling of the user-selected area on a neural implicit field with a texture template (\eg, furry hair or golden metal in Fig.~7 (c)) from a pre-captured object model.
First, we need to obtain the target UV-map of the selected area, \ie, utilizing Blender to unwrap the UV-map of the selected vertices.
Then, we select a texture template from a pre-trained NeuMesh model, \eg, a small squared patch with $\sim$10 vertices, and repeatedly fill the target UV-map with the template in a sliding-window manner.
Practically, we assign texture codes in the target vertices with interpolated codes from the template and also bind the radiance decoder to the one from the pre-trained model, as the target texture code and the template texture code do not share the same latent radiance space.
Besides, to make a smooth transition near the area boundary (\eg, naturally transiting from the edited appearance to the original appearance), for each query point that has texture codes/decoders from different sources, we fuse the color contribution from different decoders with inverse distance weighting.

\noindent\textbf{Texture painting.}
As introduced in our main paper (Sec.3.4), we propose a spatial-aware optimization to precisely transfer the painting from 2D image to 3D field, while keeping geometry and appearance of other parts unchanged.
In detail, we first shoot probing rays from the painted pixels to the mesh scaffold, and find the affected texture codes by collecting the vertices of the hit faces.
During optimization, we adopt Adam optimizer with the fixed learning rate of 0.01, and only allow these codes to be changed.
The whole texture painting optimization takes about $\sim$1 hours with 8000 iterations.

\section{More Discussions}
\label{sec:supp_discussion}

\begin{figure}[t!]
    \centering
    \includegraphics[width=0.95\linewidth, trim={0 0 0 0}, clip]{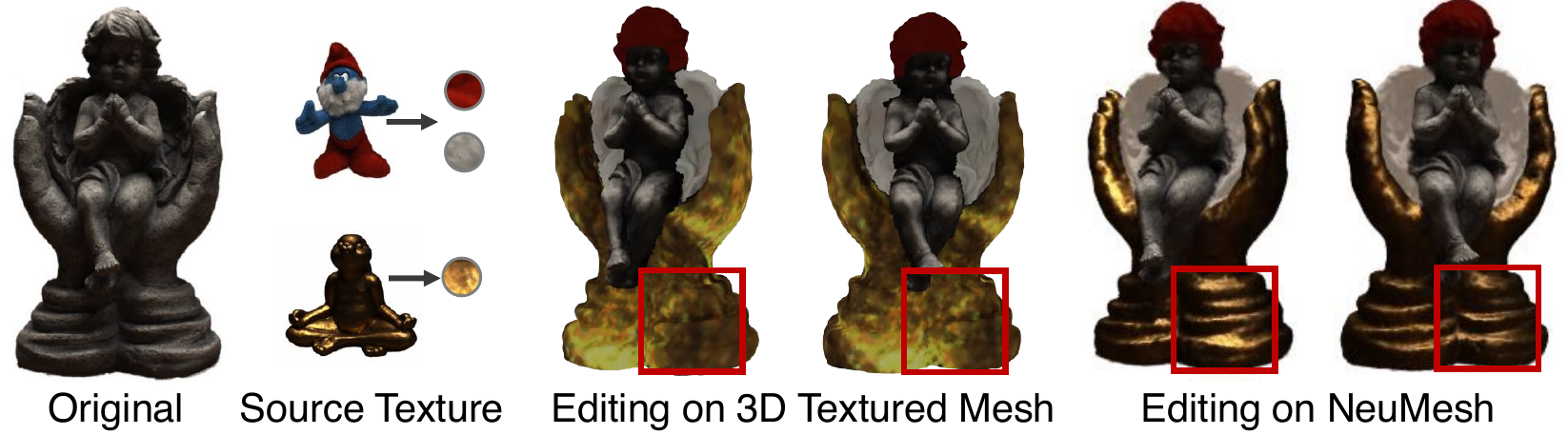}
    \label{fig:supp_traditional_mesh_edit}
    \caption{We show the comparison between textured mesh editing and our NeuMesh editing on a statue.
    This proves that direct editing texture meshes with template patterns without lighting and material property estimation cannot provide satisfactory results.
    }
\end{figure}

\noindent\textbf{Using neural implicit representation instead of traditional textured mesh.}
Neural implicit representation merits easy-reconstruction with photo-realistic volumetric rendering and view-dependent effects (\eg, shiny golden materials) on both real-world and synthetic data, and flexibility to accomplish some fine-grained editing demands (\eg, material editing or appearance variations) on the real-world scene with latent space operations.
While the rendering quality of the textured mesh is bounded by the MVS reconstruction and texturing.
It is not feasible for the textured 3D mesh to achieve such effects (see Fig.~\ref{fig:supp_traditional_mesh_edit}) without BRDF material properties and lighting estimation.

\noindent
\textbf{Using learnable signed distance.}
Unlike voxel-based~\cite{nsvf,yu2021plenoxels} or point-cloud-based~\cite{ost2021neural} methods that possess spatially scattered features, we only learn a set of `single layer' features on mesh surfaces as we want to build a surface-aligned implicit field.
Therefore, a bare code interpolation is not sufficient to coordinate the query relative position for our mesh-based representation, (\ie, the inner and outer point queries along the direction perpendicular to the surface still lack spatial distinguishability).
One plausible solution is to use a physically computed signed distance to the surface as Liu \etal~\cite{neural_actor} does, but it is not applicable for general object meshes because the geometry is not always well-defined (\eg, watertight or even predefined skinning weights) as a human-body model (SMPL)~\cite{loper2015smpl}, which confuses ray-to-mesh intersection counting and the sign of the distance might be unexpectedly reversed.
Therefore, we propose to use a learnable sign indicator to compute interpolated signed distances for spatial query points, as described in Sec.~3.1.

\noindent
\textbf{Using distillation instead of training from scratch.}
As explained in Sec.~3.2, we exploit the teacher NeuS model with distillation and fine-tuning training scheme instead of training from scratch.
The teacher NeuS model serves two purposes: 
\textbf{1)} it provides an SDF field where we could extract a mesh scaffold;
\textbf{2)} the locally embedded geometry and appearance features in our model facilitate region-based editing but may lead the training to fall into a local minimum (as shown in our ablation studies), and the use of distillation helps to alleviate such training issue.

\begin{figure}[t!]
    \centering
    \includegraphics[width=1.0\linewidth, trim={0 0 0 0}, clip]{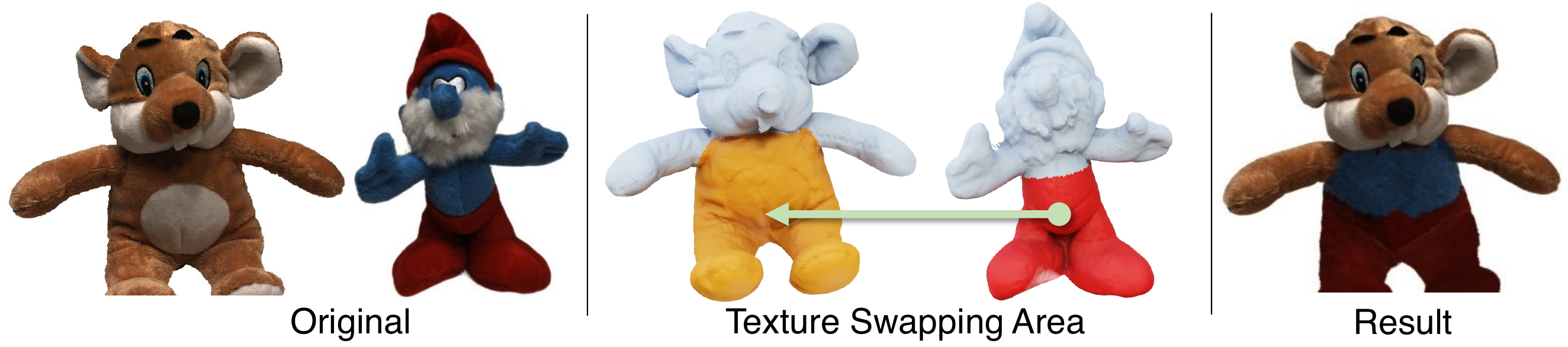}
    \label{fig:swap_diff_geo}
    \caption{Texture swapping with different geometry.
    }
\end{figure}

\noindent
\textbf{Texture swapping with different geometry/topology.}
Our method can be applied to objects with a moderate geometry difference (see Fig.~\ref{fig:swap_diff_geo}).
If there is a significant topology difference between two objects, we suggest using texture filling (Sec.~3.4 (2)) that swaps textures in UV spaces regardless of object geometries.

\noindent\textbf{Limitations for real-world applications.}
Currently, our rendering speed (about 30s for each view) is bounded by the intensive network queries and nearest neighboring searching operations.
When deploying to real-world applications, we might consider accelerating the inference speed to fulfill the real-time rendering demand
with recently proposed coefficient caching techniques~\cite{yu2021plenoxels,garbin2021fastnerf}
, multiresolution hash encoding~\cite{muller2022instant} or MVS priors~\cite{mvsnerf}.
Besides, we rely on 3D modeling software to select vertices for the region of interest, which can be replaced with some semantic annotation approaches~\cite{wang2018sgpn} to facilitate broaden users.

\noindent\textbf{Relation to point-based methods.}
From a high-level perspective, both ours and point-based methods can be regarded as building upon local feature-based representations, while the main differences include:  
\textbf{1)} Our model encodes features on mesh vertices, so we can easily deform objects with a mesh proxy or modify textures through a UV space.
Point-based methods use scattered point features, so it is non-trivial to perform mesh-based editing like ours, \ie, each point that is projected at the pixel (NPBG) or lying nearby ray samples (Point-NeRF) would contribute to the appearance, making it hard to distinguish which point features should be edited.
\textbf{2)} We embed surface normal (similar to IDR/NeuS) to realize view-dependent effects of texture filling, which cannot be directly inherited by point-based methods.

\section{More Experiment Results}
\label{sec:expr}

\begin{figure}[t!]
    \centering
    \includegraphics[width=1.0\linewidth, trim={0 0 0 0}, clip]{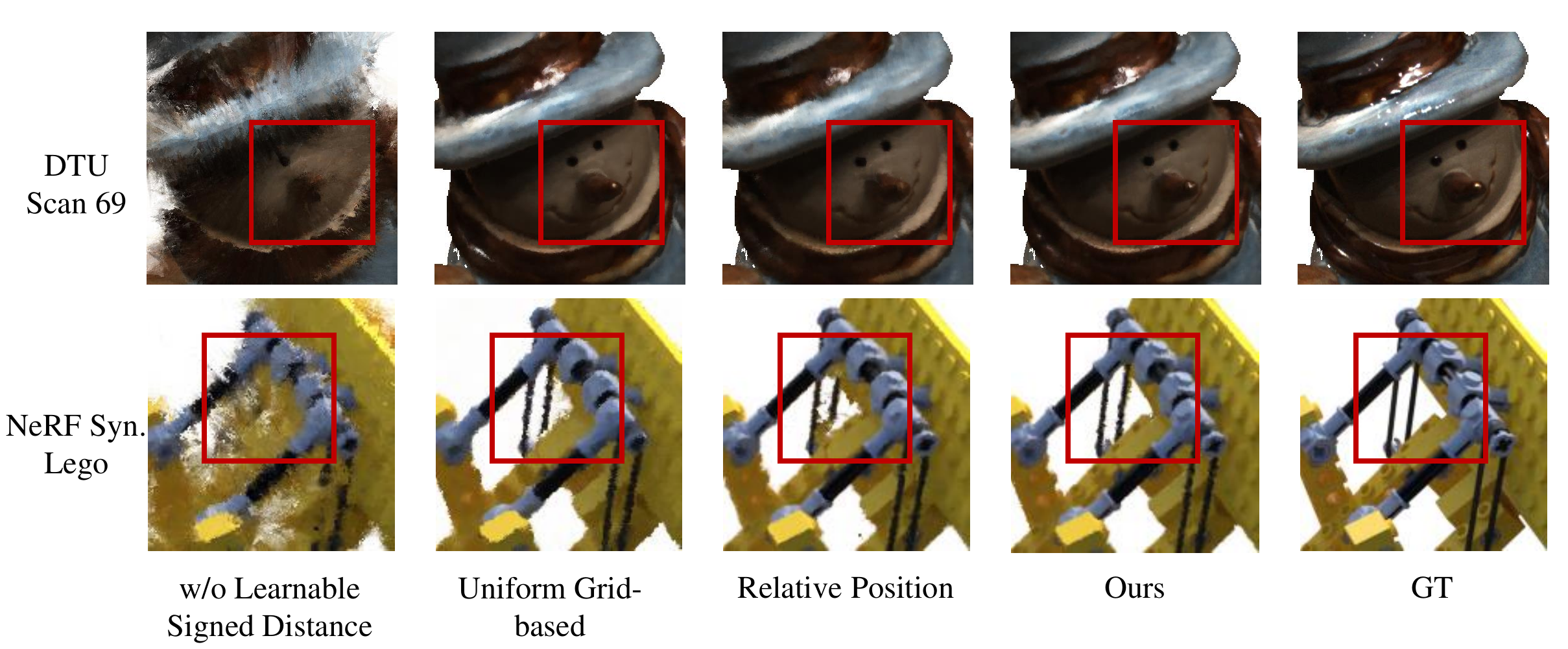}
    \caption{
    \textbf{
    We present visual comparison to alternative designs.}
    }
    \label{fig:supp_ablation}
\end{figure}
\begin{table}[t!]
\centering
\resizebox{1.0\linewidth}{!}{
\tabcolsep 10pt
\begin{tabular}{lcccccc}
\toprule
\multicolumn{1}{c}{\multirow{2}{*}{Config.}} & \multicolumn{3}{c}{DTU 69} & \multicolumn{3}{c}{NeRF 360$^{\circ}$ Synthetic Lego} \\ \cmidrule(lr){2-4} \cmidrule(lr){5-7}  
\multicolumn{1}{c}{} & \multicolumn{1}{l}{PSNR $\uparrow$} & \multicolumn{1}{l}{SSIM $\uparrow$} & \multicolumn{1}{l}{LPIPS $\downarrow$} & \multicolumn{1}{l}{PSNR $\uparrow$} & \multicolumn{1}{l}{SSIM $\uparrow$} & \multicolumn{1}{l}{LPIPS $\downarrow$} \\ \hline
w/o Learnable Signed Distance & 23.622 & 0.865 & 0.210 & 20.827 & 0.827 & 0.240 \\
Uniform Grid & 26.931 & 0.943 & 0.117 & 25.866 & 0.898 & 0.094 \\
Relative Position & 26.308 & 0.937 & 0.128 & 27.270 & 0.918 & 0.055 \\
Ours & \textbf{27.254} & \textbf{0.946} & \textbf{0.113} & \textbf{27.881} & \textbf{0.926} & \textbf{0.046} \\
\bottomrule
\end{tabular}
}
\caption{
We perform more experiments to analyze the model design with DTU Scan 69 and NeRF 360$^\circ$ Synthetic Lego.
}
\label{tab:supp_ablation}
\end{table}

\noindent\textbf{Rendering quality comparison.}
We present more results of rendering quality comparison in Fig.~\ref{fig:supp_render_compare}.
It is clear that our method renders more details than other competitors, especially when reconstructing with complex shapes and textures (\eg, the roof at DTU Scan 37, and the metal grids at NeRF-Synthetic Mic in Fig~\ref{fig:supp_render_compare}).

\begin{figure}[t!]
    \centering
    \includegraphics[width=1.0\linewidth, trim={0 0 0 0}, clip]{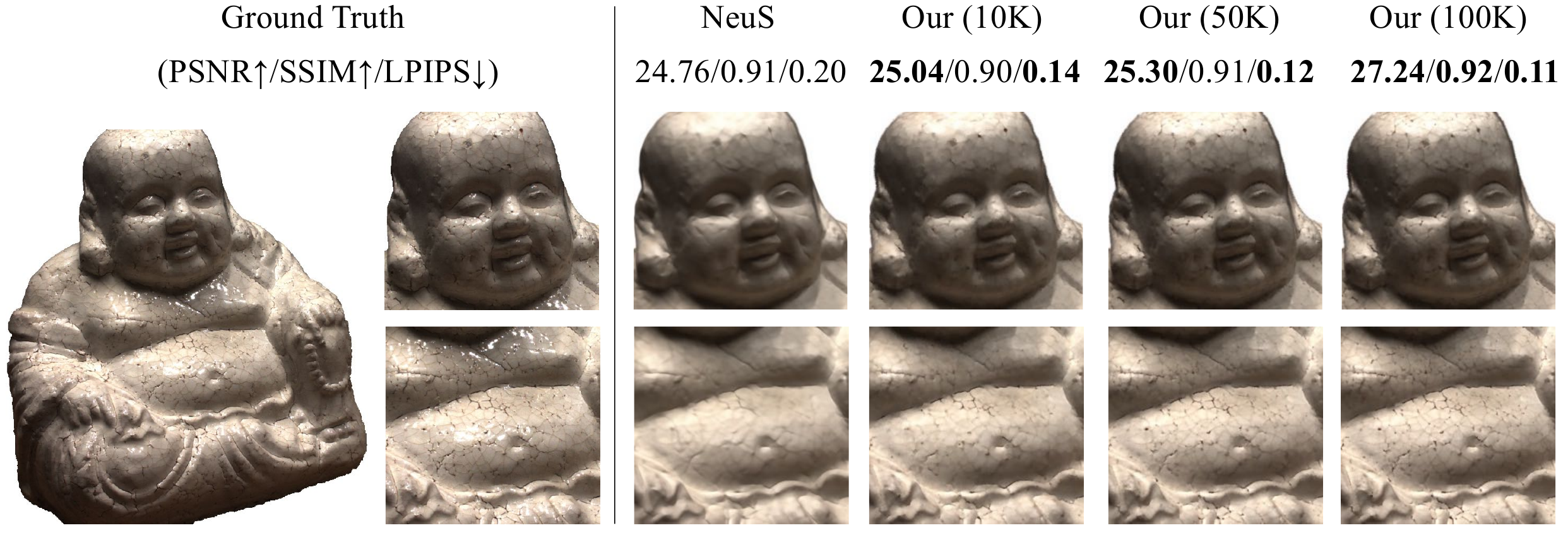}
    \caption{
    We analyze the impact of vertex numbers on the rendering quality by training with 10K / 50K / 100K vertices.
    }
    \label{fig:ablation_vertex_num}
\end{figure}

\noindent\textbf{Rendering quality with varying mesh vertex numbers.}
We analyze the impact of varying mesh vertex numbers on rendering quality.
Specifically, we train on DTU Scan 114 with 3 sets of mesh vertices (10K, 50K, and 100K).
As shown in Fig.~\ref{fig:ablation_vertex_num}, the metric quality of rendered images are slightly affected when decreasing vertex numbers, but still outperform NeuS even with only 10K vertices, which demonstrate the robustness and advantages of our representation.

\noindent\textbf{Learnable signed distance.}
We report the training results without learnable signed distance as network input in Fig.~\ref{fig:supp_ablation} (first column) and Tab.~\ref{tab:supp_ablation} (first row), which proves the necessity of this design in our mesh-based representation, as it complements spatial distinguishability on the direction perpendicular to the surface (Sec.~3.1).

\noindent\textbf{Mesh-based representation vs. uniform grid-based representation.}
We first compare our `single layer' mesh-based representation with a uniform grid-based representation (\ie, similar to NSVF~\cite{nsvf} or Plenoxel~\cite{yu2021plenoxels}).
Specifically, we thicken the mesh vertices to uniform grids, so the interpolated codes can be fully aware of the spatial coordinates, and the signed distance can be omitted.
Note that this also loses some flexibility for fine-grained editing.
As shown in Fig.~\ref{fig:supp_ablation} (second column) and Tab.~\ref{tab:supp_ablation} (second row), even with only a single slice of spatial features, our method shows on par visual quality with these uniform grid-based representation, but enables the functionalities of geometry and texture editing.

\noindent\textbf{Learnable signed distance vs. relative position.}
We then compare the encoding of our learnable signed distance with an alternative design, \ie, relative position encoding from PointNet~\cite{qi2017pointnet}.
Specifically, for each query point, we first concatenate codes (from nearby vertices) and relative coordinate offsets (from query to vertex), and encode with a shallow MLP (with 2 hidden layers and 64 hidden sizes).
Then, we use the same inverse distance weighting to obtain the final interpolated embedding for the query.
As demonstrated in Fig.~\ref{fig:supp_ablation} (third column) and Tab.~\ref{tab:supp_ablation} (third row), our learnable signed distance encoding shows better rendering quality when incorporated with such `single layer' surface features and is a better choice for mesh-based representation.

\begin{figure}[t!]
    \centering
    \includegraphics[width=0.8\linewidth, trim={0 0 0 0}, clip]{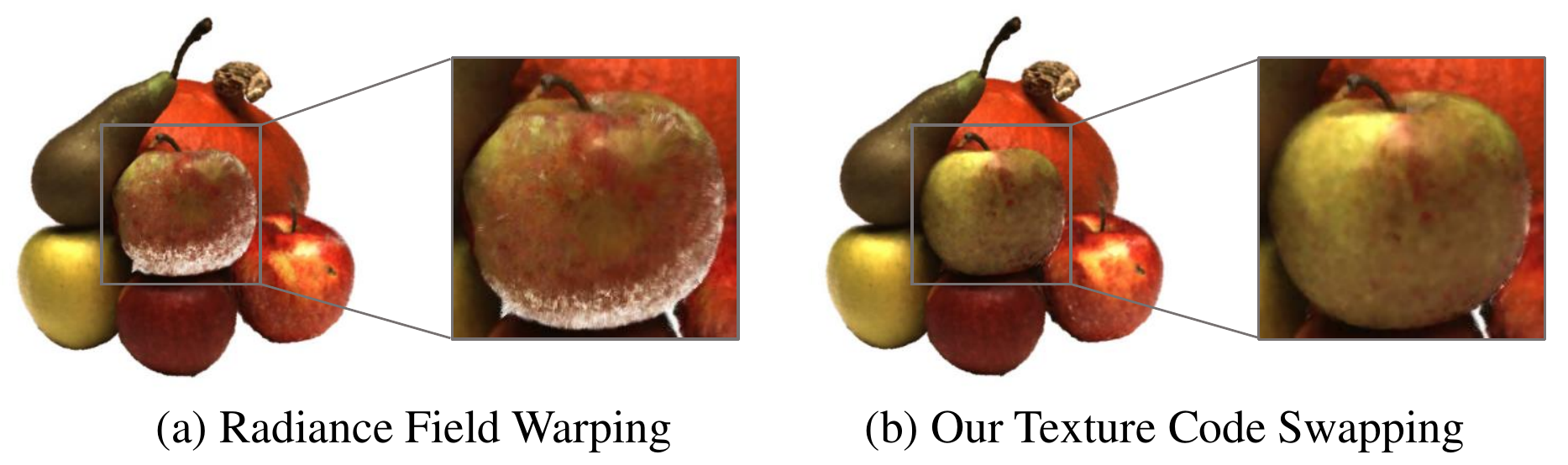}
    \caption{
    \textbf{We show the comparison of our texture editing to the field warping.}
    }
    \label{fig:supp_edit_comparsion}
\end{figure}

\noindent\textbf{Our texture editing vs. radiance field warping.}
One possible workaround of texture editing is to warp the radiance field from the original space to the aligned space according to the non-rigid mesh alignment (Sec.~3.4).
So, we compare our code updating based texture editing with such na\"ive radiance field warping on DTU Scan 63.
As shown in Fig.~\ref{fig:supp_edit_comparsion},
the rendered apple of the na\"ive approach contains noticeable artifacts, while our editing result is visually much more natural.
We believe that this is mainly due to the fact that the warped texture field might not be compatible with the geometry (SDF field), which leads to spatial misalignment (\eg, SDF field is close to the surface while radiance field is not) during the volume rendering process and produces erroneous color.
In contrast, since our method exchanges textures through code swapping, the edited texture field is tightly fit to the geometry, which yields a better rendering quality.

\begin{figure}[t!]
    \centering
    \includegraphics[width=1.0\linewidth, trim={0 0 0 0}, clip]{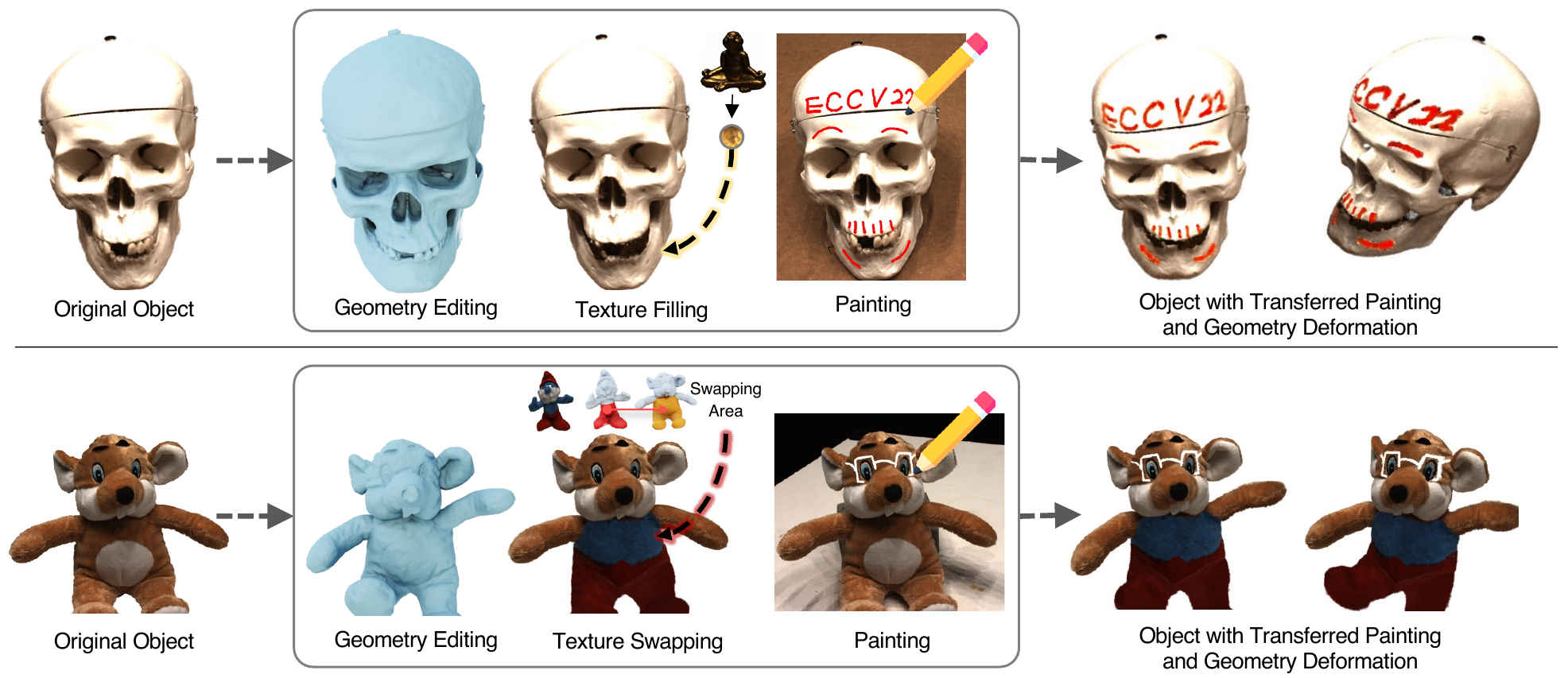}
    \caption{
    We show examples of hybrid object editing by combining multiple editing operations.
    }
    \label{fig:hybrid_edit}
\end{figure}

\noindent\textbf{Hybrid object editing.}
\noindent
To demonstrate the editing flexibility of our method, we show examples of hybrid object editing in Fig.~\ref{fig:hybrid_edit} by combining geometry/texture editing operations, which sheds light on integrating our representation into modern 3D modeling workflow.

\begin{figure}[t!]
    \centering
    \includegraphics[width=1.0\linewidth, trim={0 0 0 0}, clip]{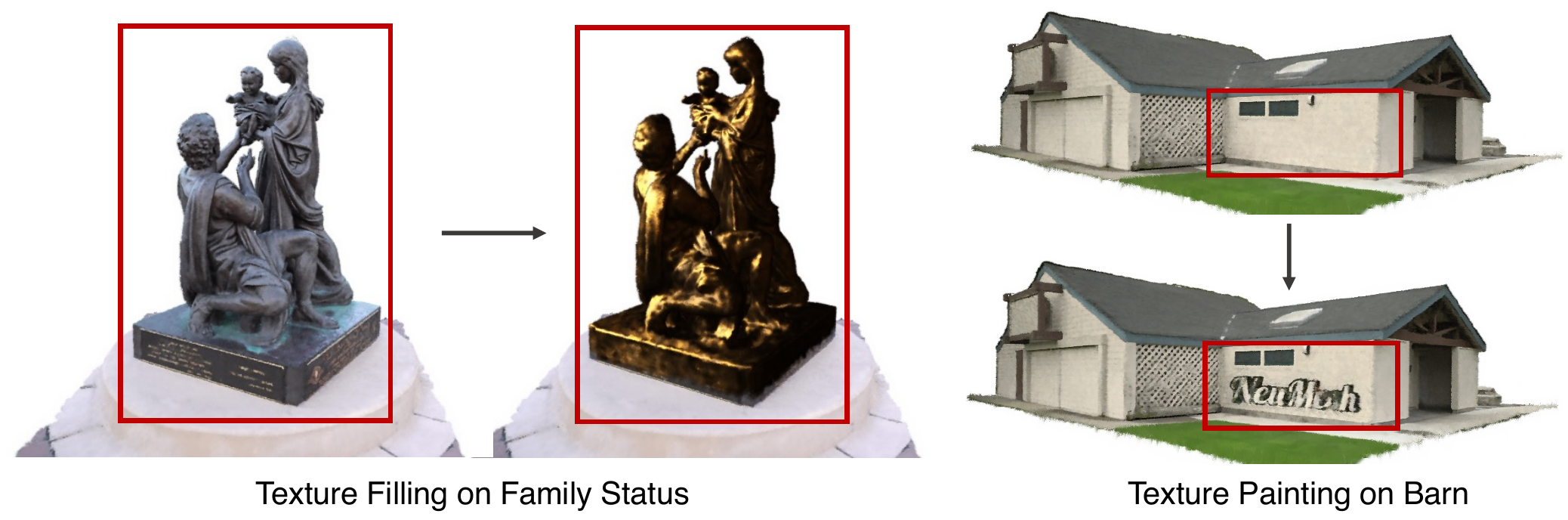}
    \caption{Texture editing of large-scale scenes on the Tanks\&Temple~\cite{Knapitsch2017} dataset.}
    \label{fig:supp_large_scale}
\end{figure}

\begin{figure}[t!]
    \centering
    \includegraphics[width=1.0\linewidth, trim={0 0 0 0}, clip]{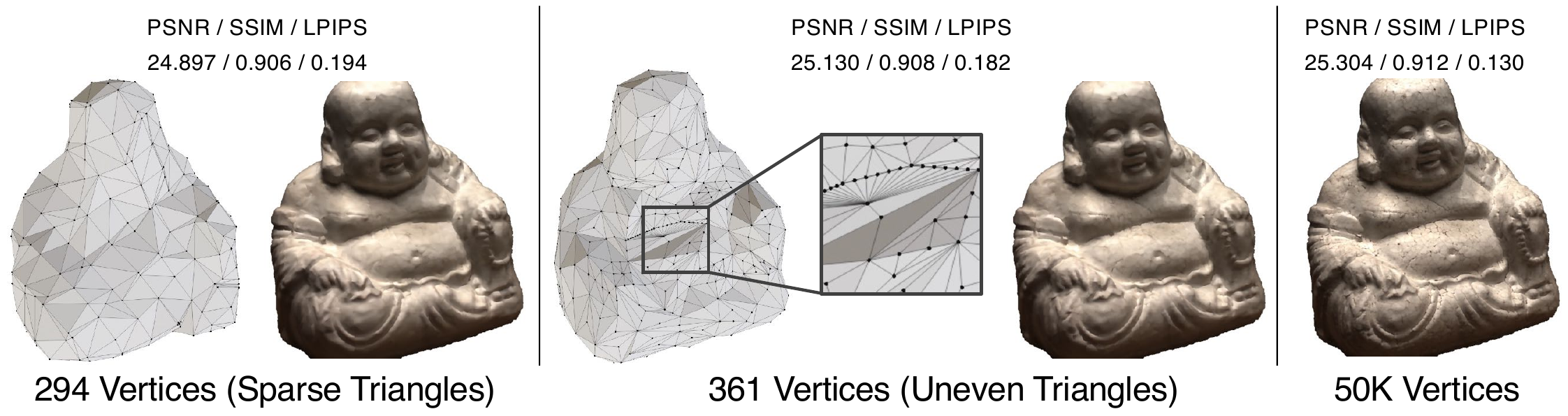}
    \caption{Rendering quality under sparse and uneven triangulation.}
    \label{fig:supp_uneven_triangle}
\end{figure}

\noindent
\textbf{Large-scale scenes.}
The modeling ability of our method depends mainly on the teacher SDF model.
As long as the scaffold mesh is available, our method can be freely scaled-up thanks to the locally embedded features.
For large scenes with complicated backgrounds, we can adopt NeRF++~\cite{nerf++}'s parameterization to handle unbounded backgrounds, or use pre-computed segmentation masks like in NSVF and IDR. 
Here we show two texture editing examples (see Fig.~\ref{fig:supp_large_scale}) on the Tanks\&Temple dataset~\cite{Knapitsch2017} with foreground segmentation provided by NSVF~\cite{nsvf}.

\noindent
\textbf{Influence of triangle quality.}
Our method can still deliver reasonable rendering quality with locally sparse/uneven triangulation (see Fig.~\ref{fig:supp_uneven_triangle}).
In fact, as the mesh scaffold is created based on the SDF from teacher NeuS, we can handily guarantee a uniformed distribution of vertices with off-the-shelf mesh regularization algorithms (\eg, isotropic remeshing by Botsch \etal~\cite{botsch2004remeshing}).

\end{document}